\PassOptionsToPackage{table}{xcolor}
\documentclass{article}
\usepackage[T1]{fontenc}
\usepackage{iclr2027_conference,times}
\usepackage{amsmath,amssymb,mathtools,booktabs,multirow,array,graphicx,xcolor}
\usepackage{microtype}
\usepackage{tabularx}
\usepackage{hyperref,url}
\usepackage{caption}
\usepackage{wrapfig}
\usepackage{placeins,float}
\hypersetup{colorlinks=true,linkcolor=teal!65!black,citecolor=teal!65!black,urlcolor=teal!65!black}
\definecolor{draftblue}{HTML}{235A75}

\newcommand{\tb}[1]{\textbf{#1}}
\newcommand{\ul}[1]{\underline{#1}}
\usepackage{multirow}

\title{What Limits Cross-Task Generalization in World Action Models?}
\author{Chen Zhao\textsuperscript{\normalfont{1,2,3$\ast$}},
  Zipeng Ye\textsuperscript{\normalfont{3$\ast$}},
  Yihan Lin\textsuperscript{\normalfont{1,2}},
  Yuchen Xie\textsuperscript{\normalfont{3}},
  Jing Zhang\textsuperscript{\normalfont{1,2$\dagger$}} \\
  \textsuperscript{1} Renmin University of China, Beijing 100872, China\\
  \textsuperscript{2} Key Laboratory of Data Engineering and KnowledgeEngineering, MOE, China\\
  \textsuperscript{3} Meituan, Beijing 100102, China\\
  \texttt{<2021201645, zhang-jing>@ruc.edu.cn}\\
}
\iclrfinalcopy
\begin{document}
\raggedbottom
\maketitle
\fancyhead{}
\begingroup\def\thefootnote{$\ast$}\footnotetext{Equal contribution.}\endgroup
\begingroup\def\thefootnote{$\dagger$}\footnotetext{Corresponding author.}\endgroup
\begin{abstract}
Cross-task generalization remains difficult for world action models (WAMs), even when their predicted futures depict plausible manipulation. This gap can be attributed directly: we make this gap directly testable by decoupling a WAM into a world model and an inverse-dynamics model (IDM), then holding the IDM fixed while replacing the source of future images. We introduce \emph{Align Once, Evaluate Everywhere} (AoE), a camera-space inverse-dynamics model pretrained on large-scale robot data, which provides this recovery interface for different future sources across tasks and environments. Through it, we find that actions recovered from recorded futures of unseen tasks execute almost as well as replayed demonstrations, whereas futures predicted by world models collapse across benchmarks. The failures are not generic: they concentrate on the gripper and interaction geometry while generation capacity is spent on task-irrelevant background. This diagnosis suggests that supervision should target exactly these regions. We therefore introduce the \emph{vision-to-action} (VITA) loss, which passes generated futures through the frozen AoE and penalizes discrepancies in the recovered actions, redirecting generation toward action-relevant detail. Across two world models and three manipulation benchmarks, this simple change raises the execution success of generated futures from near zero to 47.0\%. The same loss also transfers to WAMs and yields consistent gains in native control.
\end{abstract}
\section{Introduction}
\label{sec:introduction}

World action models (WAMs) offer a compelling premise for embodied tasks: use the generalization capabilities of world models (WMs) to imagine future scenes, then condition an action head on the latent representations of those imagined futures to generate actions~\citep{lingbotva,fastwam}. This approach achieves strong performance on in-distribution tasks and under modest distribution shifts, yet falls substantially behind in more challenging settings, such as generalization across environments or tasks. This gap is often attributed to the difficulty of predicting 3D poses or to action heads that are tightly coupled to the environment encountered during training. Recent studies have also identified problems in world model predictions of robotic scenes, including physically implausible behavior and 2D-like flat appearance~\citep{worldarena,worldsimprobe}. However, these analyses remain confined to 2D future videos that cannot be executed directly, leaving the impact of world model predictions on downstream cross-task generalization unquantified.

Evaluating these videos through an inverse dynamics model (IDM) can also be inconclusive: exposure to trajectories from downstream environments during training may allow the IDM to compensate for distorted predictions, obscuring whether the predicted futures themselves support execution. To better evaluate the executability of future videos predicted by world models, we develop \emph{Align Once, Evaluate Everywhere} (AoE). Pretrained on 9,637h of robot data, AoE uses camera-space representation to decouple action prediction from downstream environments. With little data from a downstream environment, AoE translates 2D gripper displacements and depth changes into executable poses. 
\begin{figure}
\centering
\includegraphics[width=\linewidth,height=2.45in,keepaspectratio]{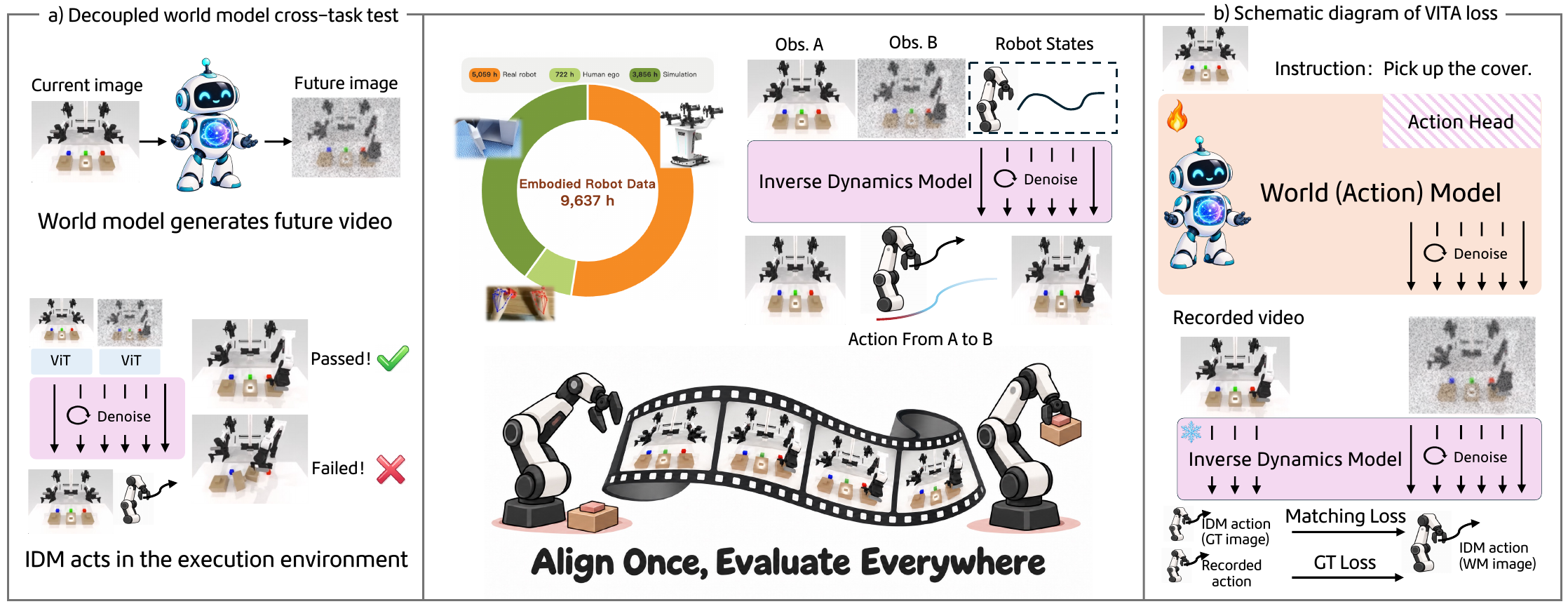}
\caption{\textbf{Overview of Align Once, Evaluate Everywhere (AoE)}. Trained on 9,637 hours of robot image–action pairs, AoE infers the action between any two frames, serving as a bridge from video to action. \emph{Left:} AoE enables evaluation of WMs: dreamed task completions in unseen environments are decoded into executable actions and assessed in the real environment. \emph{Right:} The proposed vision-to-action (VITA) loss further improves the world (action) model by aligning actions recorded and inferred from dreamed and recorded videos.}
\label{fig:overview}
\end{figure}
This enables interactive execution beyond the environments seen during pretraining, while keeping action trajectories out of training. On the same unseen benchmarks, we observe a substantial gap between recorded videos and world model predictions: actions recovered from recorded videos approach the success rate of replaying recorded actions, whereas those recovered from predicted future videos perform poorly. 2D flat appearance, physically implausible pose, and hallucinated scene substantially degrade the quality of the recovered poses~\citep{liu2026geometryaware4dvideogeneration, mi2026tcidmgroundingvideogeneration}.

To address this limitation, we introduce the \emph{vision-to-action} (VITA) loss, which uses AoE as a frozen supervisor to guide world models toward preserving the visual details needed for physically accurate gripper motion. We pass predicted future videos through AoE and use discrepancies of action-space representation to supervise action-relevant visual details~\citep{wang2026evaaligningvideoworld,feng2025vidarc}. AoE's sensitivity to gripper motion directs this supervision toward regions that matter for downstream execution~\citep{wi2023calamari,wu2020spatialactionmapsmobile}. Gradients propagate through AoE to the world model while AoE itself remains frozen. VITA is natively suitable for world action model training to supervise predicted visual futures, allowing action-grounded visual supervision to shape the representations used for native action prediction. This supervision is applied during training and does not require the model to explicitly generate future images during execution.

Our experiments examine three aspects of this approach. First, we use AoE to evaluate the cross-task executability of future videos generated by existing world models. Comparing generated futures with recorded videos reveals a substantial execution gap, highlighting a limitation in the transfer of world-model predictions to unseen tasks. Second, we evaluate VITA in both world models and world action models. VITA substantially improves the execution success of actions recovered from generated futures and improves native action prediction in LingBot-VA~\citep{lingbotva} and FastWAM~\citep{fastwam}, with particularly pronounced gains under distribution shifts~\citep{zheng2026giftguidedintermediatefeature,zhen2026actionimagesendtoendpolicy}. FastWAM benefits even though it does not explicitly generate future images during inference. Third, we study how scaling AoE's pretraining data affects its roles as both an evaluator and a supervisor. Larger-scale pretraining improves action recovery on unseen tasks and yields stronger downstream performance when the frozen AoE is used for VITA supervision.

Our contributions are threefold: (1) we introduce AoE, an inverse dynamics model that uses the camera-space action representation to support zero-shot cross-task generalization; (2) we propose VITA, an action-grounded loss that uses frozen AoE supervision to guide both WMs and WAMs to preserve visual details needed for executable 3D pose changes; and (3) we demonstrate that VITA raises the cross-task execution success of world-model-generated futures from near zero to 47.0\% and improves native action generation in WAMs under both in-distribution and out-of-distribution.

\section{Preliminaries}
\label{sec:preliminaries}

\paragraph{Forward and inverse dynamics.}\label{sec:dynamics} Let $o_t$ denote the robot's observation at time $t$, $a_t$ its action, and $\ell$ a language instruction specifying the task. A forward dynamics model (FDM) predicts the consequences of an action. Written in observation space, its prediction takes the form
\begin{equation}
    \hat{o}_{t+1} = \mathcal{F}_{\theta}(o_t, a_t),
    \label{eq:forward_dynamics}
\end{equation}
where $\theta$ denotes the model parameters and $\hat{o}_{t+1}$ is the predicted next observation. An inverse dynamics model (IDM) infers an action consistent with an observed transition:
\begin{equation}
    \hat{a}_t = \mathcal{D}_{\phi}(o_t, o_{t+1}),
    \label{eq:inverse_dynamics}
\end{equation}
where $\phi$ denotes the IDM parameters. These two mappings connect actions and observations in complementary directions: forward dynamics predicts what an action will produce, while inverse dynamics recovers an action that could produce a given transition. Applied to selected frame pairs from a video, inverse dynamics provides a way to translate visual changes into robot actions.

\paragraph{From world models to world action models.}\label{sec:wm_to_wam} In embodied tasks, a world model provides the visual-prediction component of a policy. An instruction-conditioned world model predicts a task-directed future observation from the current observation and instruction. WAMs integrate this prediction process with action generation~\citep{lingbotva,fastwam}. In attention-based architectures, an action head reads internal visual features associated with the predicted future. We summarize this interface as
\begin{equation}
    h_t = \mathcal{H}_{\theta}(o_t, \ell),
    \qquad
    \hat{a}_t = \mathcal{A}_{\psi}(o_t, \ell, h_t),
    \label{eq:coupled_wam}
\end{equation}
where $h_t$ denotes the latent features associated with future prediction, $\mathcal{H}_{\theta}$ denotes their computation within the world model, and $\mathcal{A}_{\psi}$ is the action head. Attention allows the action head to access $h_t$ during action prediction. Training this coupled system on downstream data ties the visual features to the action head that consumes them. Future imagination therefore influences control through an internal representation, and task success reflects the behavior of the combined system. This makes it difficult to directly assess how the predicted visual future itself supports execution on unseen tasks.

We introduce an explicit image-space interface between future prediction and action recovery:
\begin{equation}
    \hat{o}_{t+1} = \mathcal{W}_{\theta}(o_t, \ell),
    \qquad
    \hat{a}_t = \mathcal{D}_{\phi}(o_t, \hat{o}_{t+1}),
    \label{eq:decoupled_prediction}
\end{equation}
where $\mathcal{W}_{\theta}$ is the instruction-conditioned world model and $\mathcal{D}_{\phi}$ is an independently pretrained IDM. We first evaluate the IDM on recorded transitions from unseen tasks. We then replace the recorded future observations with world-model predictions while keeping the IDM frozen. This shared action-recovery interface allows us to evaluate IDM transfer and the executability of generated futures separately under the same execution protocol.
\section{Method}
\label{sec:method}

We introduce AoE, an inverse dynamics model pretrained on 9,637 hours of robot image–action pairs, and use it as a frozen interface for evaluating and improving WMs and WAMs for embodied tasks. Section~\ref{sec:aoe} presents AoE's overview and our protocol for evaluating the executability of future videos generated for unseen tasks. Section~\ref{sec:video_action_loss} introduces the AoE-based vision-to-action (VITA) loss and its integration into world-model and WAM training. Section~\ref{sec:supervision_analysis} analyzes how this loss guides future prediction and improves downstream execution.

\begin{figure}[t]
\centering
\includegraphics[width=\linewidth,height=3.85in,keepaspectratio]{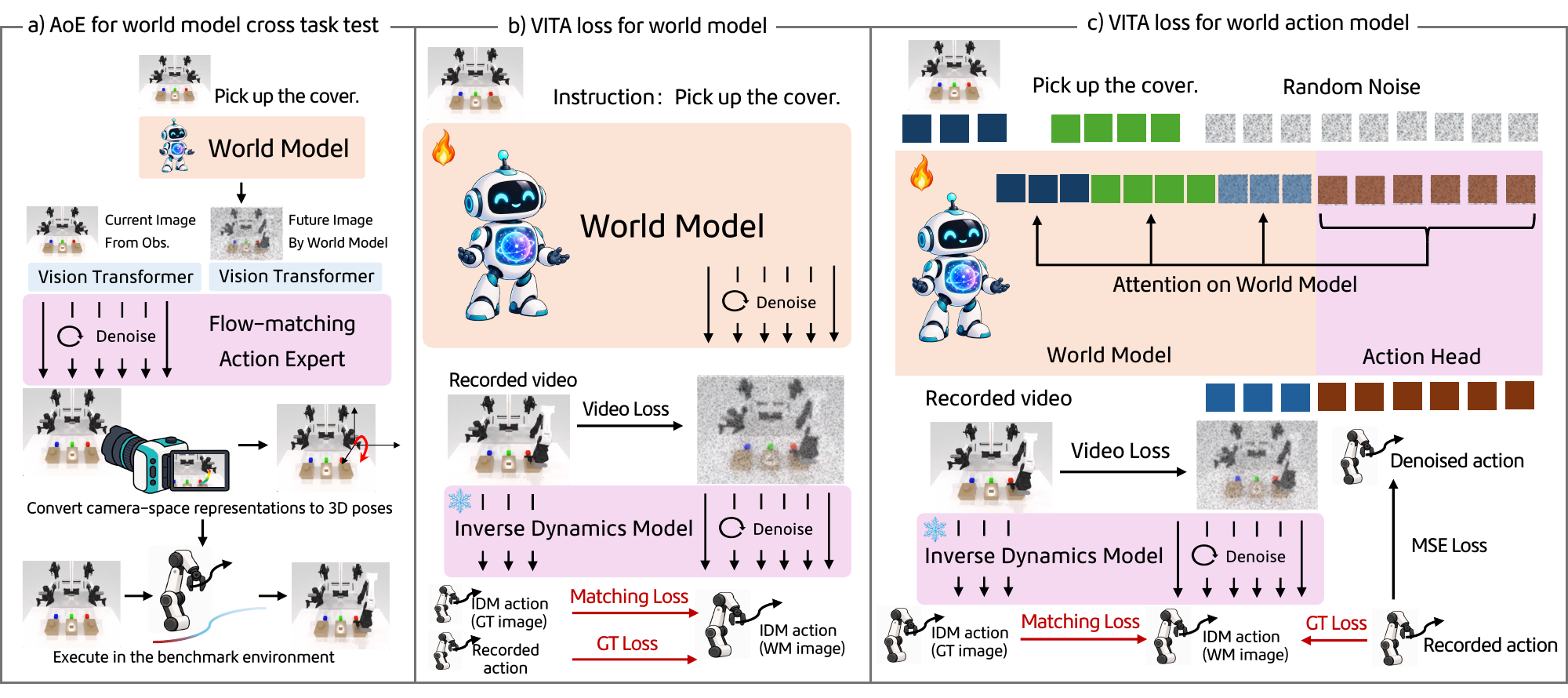}
\caption{\textbf{Cross-task generalization evaluation and the VITA loss.} \emph{Left:} Future videos generated by the WM in unseen environments are fed into AoE frame-by-frame to obtain camera-space representations, which are translated into 3D poses and executed in the real environment for evaluation. \emph{Middle:} Dreamed future frames are decoded by AoE into actions, and the WM is updated by minimizing their discrepancies against actions inferred from recorded videos as well as the ground-truth actions. \emph{Right:} The vision-to-action loss is transferred to WAM. By encouraging the future video to focus on task-critical regions, the native action generation of WAM is improved under cross-task generalization.}
\label{fig:method}
\end{figure}

\subsection{Camera-Space Inverse Dynamics for Cross-Task Evaluation}
\label{sec:aoe}

An IDM infers actions that connect a current observation $o_t$ to a future observation $o_{t+i}$. Predicting pose-based actions directly in environment coordinates ties this mapping to the coordinate convention of the training environment: the same end-effector motion can correspond to different action values under different world or robot-base frames. The action head consequently learns both the motion implied by the images and an environment-specific coordinate mapping.

We separate these two operations with a camera-space action representation. For arm $r \in \{\mathcal{L},\mathcal{R}\}$, AoE predicts
\begin{equation}
    \hat{a}^{\,r}_{t,t+i}
    =
    \mathcal{D}^{r}_{\phi}(o_t,o_{t+i};s_t)
    \in \mathbb{R}^{d_a},
    \label{eq:aoe_prediction}
\end{equation}
where $s_t$ is the current robot state, and $d_a$ denotes the dimension of the camera-space action target. AoE summarizes each arm's motion $\hat{a}^{\,r}_{t,t+i}$ by its projected start and end positions, relative depth change, orientation change, and gripper command from timestamp $t$ to $t+i$. This compact target captures the transition between the two observations and provides a common prediction space across environments. The target encoding and its conversion into executable poses using camera calibration are specified in Appendix~\ref{app:camera_action}; the complete architecture and token routing are given in Appendix~\ref{app:probe_details}.

Through large-scale pretraining on robot interaction data, AoE learns to recover end-effector motion from paired observations in this shared representation. Environment-specific coordinate conversion is handled explicitly at execution, allowing the learned image-to-action mapping to be reused across downstream environments. AoE remains frozen after pretraining.

\paragraph{Evaluating cross-task future videos.}

Given an initial observation and a task instruction, a WM generates a sequence of future observations. We apply AoE to consecutive frame pairs, recover the corresponding pose commands, and execute them to measure task success. This turns predicted videos into an execution-based test of cross-task generalization.

We first evaluate the same procedure on recorded videos from tasks excluded from AoE pretraining, establishing the action-recovery model's transfer performance. We then substitute world-model predictions while keeping AoE, the initial conditions, and the execution protocol fixed. If the generated futures preserve the information needed for control and AoE reliably recovers the corresponding actions, this pipeline supports zero-shot execution on unseen tasks. The comparison with recorded videos measures how replacing genuine visual transitions with predicted ones affects downstream success.

\subsection{Vision-to-Action Future Supervision}
\label{sec:video_action_loss}

World-model video objectives treat visual content broadly, while
downstream manipulation depends strongly on the relative geometry of
the gripper and the target object. AoE provides an action-grounded
training signal for this geometry. We use the pretrained AoE model as a
frozen supervisor during world-model training, so that the generated
future is evaluated according to the action information it conveys.

Given the current observation $o_t$, let $o_{t+i}$ denote the recorded
future observation and let $\hat{o}_{t+i}$ denote the corresponding
world-model prediction. We pass both observation pairs through AoE
under the same robot state, noisy action, and flow time. Let
$f_{\phi}(o_t,o)$ denote AoE's action-flow prediction conditioned on the
observation pair $(o_t,o)$, and let $\mathbf{v}^{\star}_{t,i}$ denote
the ground-truth action-flow target. The core objectives are
\begin{equation}
\begin{aligned}
    \mathcal{L}_{\mathrm{match}}
    =
    \mathbb{E}
    \left[
        \left\|
            f_{\phi}(o_t,\hat{o}_{t+i})
            -
            f_{\phi}(o_t,o_{t+i})
        \right\|_2^2
    \right],
    \quad
    \mathcal{L}_{\mathrm{GT}}
    =
    \mathbb{E}
    \left[
        \left\|
            f_{\phi}(o_t,\hat{o}_{t+i})
            -
            \mathbf{v}^{\star}_{t,i}
        \right\|_2^2
    \right],
\end{aligned}
\label{eq:video_action_losses}
\end{equation}
where the expectation averages over training transitions and the
shared action-noising conditions. We refer to their sum,
$\mathcal{L}_{\mathrm{VITA}}
=
\mathcal{L}_{\mathrm{match}}
+
\mathcal{L}_{\mathrm{GT}}$,
as the VITA loss.

The matching term transfers action-relevant information from the
recorded future, while the ground-truth (GT) term anchors this information to the
action that produced the transition. We add $\mathcal{L}_{\mathrm{VITA}}$
to the native world-model objective during cross-task training.
Gradients are propagated through the generated-image branch while AoE
remains frozen, allowing the WM to improve the visual details
that affect action recovery. The flow-target construction and
implementation details are provided in Appendices~\ref{app:video_action_loss} and~\ref{app:loss_details}.

\subsection{How vision-to-action Supervision Shapes Prediction and Control}
\label{sec:supervision_analysis}

We analyze how the VITA loss changes future prediction and
how these changes reach the native action head.
The expressions below condition on one transition and average over the shared AoE noising conditions introduced in
Section~\ref{sec:video_action_loss}.
Detailed conditions and proofs are provided in
Appendix~\ref{app:supervision_analysis}.

\paragraph{Supervising action-relevant visual details.}
The matching and GT losses jointly guide AoE's prediction toward
the average of the recorded-future prediction and the ground-truth
action-flow target:
\begin{equation}
\begin{aligned}
    \bar{\mathbf v}
    &=
    \frac{
        f_\phi(o_t,o_{t+i})+\mathbf v^\star_{t,i}
    }{2},
    \quad
    \mathcal L_{\mathrm{VITA}}
    =
    \mathcal L_{\mathrm{match}}+\mathcal L_{\mathrm{GT}}
    =
    2\mathbb E
    \left[
        \left\|
            f_\phi(o_t,\hat o_{t+i})-\bar{\mathbf v}
        \right\|_2^2
    \right]+C,
    \label{eq:analysis_combined_target}
\end{aligned}
\end{equation}
where $C$ is constant with respect to the WM.
Let $J_\phi(o)=\partial f_\phi(o_t,o)/\partial o$.
For a small visual perturbation $\Delta o$,
\begin{equation}
    \mathcal L_{\mathrm{match}}(o_{t+i}+\Delta o)
    =
    \Delta o^\top
    \mathbb E
    \left[
        J_\phi(o_{t+i})^\top J_\phi(o_{t+i})
    \right]
    \Delta o
    +O(\|\Delta o\|_2^3).
    \label{eq:analysis_visual_weighting}
\end{equation}
Thus visual errors receive stronger penalties when they have a
larger effect on action recovery. Appendix~\ref{app:visualization} visualizes AoE's
attention to end-effector regions.

\paragraph{From future images to native actions.}
Let $h$ denote the shared conditioning features used for future-observation
and native-action prediction, and let $\theta$ denote the parameters
that produce them.
Applying the chain rule to Equation~\eqref{eq:analysis_combined_target}
propagates the action-flow discrepancy through AoE and the predicted
observation into $h$, then into $\theta$. This updates the conditioning
available to the native action head while AoE remains frozen:
\begin{equation}
\begin{aligned}
    \nabla_h\mathcal L_{\mathrm{VITA}}
    &=
    4\mathbb E\left[
        \left(
            \frac{\partial\hat o_{t+i}}{\partial h}
        \right)^\top
        J_\phi(\hat o_{t+i})^\top
        \left(
            f_\phi(o_t,\hat o_{t+i})-\bar{\mathbf v}
        \right)
    \right],
    \nabla_\theta\mathcal L_{\mathrm{VITA}}
    =
    \left(
        \frac{\partial h}{\partial\theta}
    \right)^\top
    \nabla_h\mathcal L_{\mathrm{VITA}}.
\end{aligned}
\label{eq:analysis_feature_gradient}
\end{equation}
The action head continues to learn through the native objective
on these updated features.

\paragraph{A local action-error bound.}
Suppose reference features $h^\star$ reproduce the recorded future
and AoE captures the feature directions used by the action head.
Under the local smoothness conditions in
Appendix~\ref{app:supervision_analysis},
\begin{equation}
    \left\|
        \hat{\mathbf a}-\mathbf a^\star
    \right\|_2
    \le
    \kappa
    \left(
        \underbrace{
            \sqrt{\frac{\mathcal L_{\mathrm{VITA}}}{2}}
        }_{\text{VITA fitting term}}
        +
        \underbrace{
            \frac{\epsilon_{\mathrm{AoE}}}{2}
        }_{\text{teacher error term}}
    \right)
    +
    \underbrace{
        C_h\|h-h^\star\|_2^2
    }_{\text{local nonlinear remainder}}+\underbrace{
        \epsilon_{\mathrm{head}}
    }_{\text{reference head error}},
    \label{eq:analysis_native_bound}
\end{equation}
where $\hat{\mathbf a}$ is the native action prediction,
$\mathbf a^\star$ is its target,
$\epsilon_{\mathrm{head}}$ is the head's reference error,
$\epsilon_{\mathrm{AoE}}$ is the teacher's action-flow error on
the recorded transition, $\kappa$ relates their feature
sensitivities, and $C_h$ bounds local curvature.
The bound connects accurate AoE pretraining, auxiliary fitting
of generated futures, and native action learning through their
respective error terms.
\section{Experiments}
\label{sec:experiments}
We conduct extensive experiments on both the decoupled WM + IDM
architecture and WAMs, aiming to answer the following
questions:

\textbf{Q1:} Do world-model-generated future videos support execution on
unseen tasks? (Section~\ref{sec:execution_gap}).

\textbf{Q2:} Can the proposed VITA loss help world models achieve cross-task generalization, and can it transfer to WAMs, improving their native action inference? (Section~\ref{sec:vita_results}).

\textbf{Q3:} How does AoE's pretraining data scale affect its roles as evaluator and teacher? (Section~\ref{sec:aoe_scaling}).

\textbf{Q4:} How do AoE’s action representation and the components of VITA affect performance? (Section~\ref{sec:ablations}).
\paragraph{Setup.} We experiment with two world models,
Cosmos~\citep{cosmos} and WAN2.2~\citep{wan,wan22}, and two WAMs,
LingBot-VA~\citep{lingbotva} and FastWAM~\citep{fastwam}. For cross-task generalization, models are trained on the LIBERO-90 training split with the test tasks held out, and evaluated in the standard LIBERO~\citep{libero}. For RoboTwin 2.0~\citep{robotwin2}, we adopt two-fold cross-validation: 25 tasks are drawn from the clean and randomized settings for training, and the remaining 25 tasks are used for testing. For cross-environment generalization and WAM evaluation, models are trained on LIBERO and evaluated in LIBERO-Plus~\citep{liberoplus} environments. RoboTwin clean$\to$clean+random and clean+random$\to$clean+random are trained on clean-only and clean+random data, respectively; all other settings are kept identical. Detailed AoE, WM, and WAM training settings are provided in Appendices~\ref{app:training_details} and~\ref{app:full_results}.
\begin{table*}[t]
\centering
\caption{\textbf{Cross-task execution on LIBERO and RoboTwin 2.0.}
World models are trained on LIBERO-90 after excluding evaluation-task identities and RoboTwin with two-fold task-level cross-validation. Zero-shot rows use world-model-generated future videos, while Dataset image replay+AoE uses recorded dataset observations and Dataset action replay directly replays action trajectories from the dataset. The checkmark indicates that the frozen AoE model is used to recover actions from images.}
\label{tab:future_video_execution_gap}
\renewcommand{\arraystretch}{1.15}
\resizebox{\linewidth}{!}{
\begin{tabular}{l | c c c || c c c c | c || c c}
\toprule
&
&
&
& \multicolumn{5}{c||}{\textbf{LIBERO (\%)}}
& \multicolumn{2}{c}{\textbf{RoboTwin (\%)}} \\
\cmidrule(lr){5-9}
\cmidrule(lr){10-11}
\textbf{Model}& \textbf{IDM action}& \textbf{Source}& \textbf{Inference}&
\textbf{Spatial}
& \textbf{Object}
& \textbf{Goal}
& \textbf{Long}
& \textbf{Avg.}
& \textbf{Easy}
& \textbf{Hard} \\
\midrule

Dataset action replay
& $\times$
& Replay
& --
& \tb{80.4}
& \tb{88.8}
& \tb{62.0}
& \tb{57.8}
& \tb{72.3}
& \tb{66.7}
& \tb{70.0} \\

\rowcolor{blue!4}
Dataset image replay + AoE
& \checkmark
& Replay
& --
& \ul{71.6}
& \ul{74.0}
& \ul{53.0}
& \ul{29.2}
& \ul{57.0}
& \ul{65.3}
& \ul{60.7} \\

\midrule

\multirow{2}{*}{Cosmos + AoE}
& \checkmark
& Zero-shot
& pass@1
& 0.4
& 0.6
& 0.4
& 0.0
& 0.4
& 1.6
& 1.1 \\

& \checkmark
& Zero-shot
& pass@50
& 5.2
& 3.8
& 4.0
& 10.6
& 5.9
& 5.3
& 4.7 \\

\multirow{2}{*}{WAN2.2 + AoE}
& \checkmark
& Zero-shot
& pass@1
& 0.2
& 0.8
& 0.2
& 0.0
& 0.3
& 1.9
& 1.3 \\

& \checkmark
& Zero-shot
& pass@50
& 4.4
& 5.2
& 4.8
& 3.8
& 4.6
& 6.6
& 5.8 \\

\bottomrule
\end{tabular}
}
\end{table*}

\begin{table*}[t]
\centering
\caption{\textbf{Cross-environment execution on LIBERO-Plus.}
All world models are trained on LIBERO and evaluated on LIBERO-Plus.
Zero-shot rows use world-model-generated future videos, while Dataset image replay+AoE uses recorded dataset observations and Dataset action replay directly replays action trajectories from the dataset. The checkmark indicates that the frozen AoE model is used to recover actions from images.}
\label{tab:libero_plus_execution_gap}
\renewcommand{\arraystretch}{1.15}
\resizebox{\linewidth}{!}{
\begin{tabular}{l | c c c || c c c c c c c | c}
\toprule
\textbf{Model}
& \textbf{IDM action}
& \textbf{Source}
& \textbf{Inference}
& \textbf{Background}
& \textbf{Light}
& \textbf{Language}
& \textbf{Noise}
& \textbf{Layout}
& \textbf{Camera}
& \textbf{Robot Init.}
& \textbf{Overall} \\
\midrule

Dataset action replay
& $\times$
& Replay
& --
& \tb{71.9}
& \tb{75.3}
& \tb{78.0}
& \tb{83.8}
& \tb{78.8}
& \tb{67.7}
& --
& \tb{76.2} \\

\rowcolor{blue!4}
Dataset image replay + AoE
& \checkmark
& Replay
& --
& \ul{60.7}
& \ul{60.3}
& \ul{59.6}
& 0.0
& \ul{53.6}
& \ul{57.0}
& --
& \ul{47.0} \\

\midrule

\multirow{2}{*}{Cosmos + AoE}
& \checkmark
& Zero-shot
& pass@1
& 1.1
& 0.4
& 0.5
& 0.0
& 0.1
& 0.4
& 0.3
& 0.4 \\

& \checkmark
& Zero-shot
& pass@50
& 8.7
& 2.8
& 4.2
& 0.2
& 2.7
& 5.4
& \ul{2.1}
& 3.5 \\

\multirow{2}{*}{WAN2.2 + AoE}
& \checkmark
& Zero-shot
& pass@1
& 0.9
& 0.5
& 0.5
& 0.0
& 0.5
& 0.7
& 0.3
& 0.5 \\

& \checkmark
& Zero-shot
& pass@50
& 5.7
& 3.8
& 5.1
& \ul{0.4}
& 3.0
& 6.4
& \tb{2.5}
& 3.8 \\

\bottomrule
\end{tabular}
}
\end{table*}

\subsection{How Executable Are World Model Futures on Unseen Tasks?}
\label{sec:execution_gap}

To evaluate the executability of world-model-generated future videos across tasks and environments, we generate future videos with Cosmos~\citep{cosmos} and WAN2.2~\citep{wan,wan22} from held-out initial observations and instructions, and recover actions with the frozen AoE.
Table~\ref{tab:future_video_execution_gap} shows that the evaluation itself
is sound: on recorded videos, AoE recovers most of the replay performance,
reaching 57.0\% average success on LIBERO against 72.3\% for replaying
recorded actions, with RoboTwin following the same pattern. Generated
futures, however, collapse. Both world models succeed on at most 0.4\% of
LIBERO trials under pass@1. The gap persists under cross-environment evaluation
(Table~\ref{tab:libero_plus_execution_gap}), where recorded images still
reach 47.0\% overall on LIBERO-Plus while generated futures stay under 4\%.
Inspecting the failed generations reveals why: both models spend their
capacity on task-irrelevant background, while the gripper is generated with flat appearance,
physically implausible finger configurations, and hallucinated content that
contradicts the observed scene, and these defects translate directly into
erroneous recovered poses.


\begin{table*}[t]
\centering
\caption{\textbf{Effect of VITA supervision on cross-task future-video
executability.}
Zero-shot rows use future videos generated by the corresponding world
model, while the replay row uses recorded dataset observations.
Rows with ``+VITA'' use the additional action-grounded supervision.
The checkmark indicates that the AoE is used to recover
actions from images. \textcolor{green!50!black}{Green}/\textcolor{red}{Red} numbers show the \textcolor{green!50!black}{improvement}/\textcolor{red}{degradation} over baseline.}
\label{tab:vita_future_video_execution}
\renewcommand{\arraystretch}{1.15}
\resizebox{0.85\linewidth}{!}{
\newcolumntype{R}{>{\hspace{0em}}c<{\hspace{0.575em}}}
\begin{tabular}{l | c R || c c c c | c || c c}
\toprule
&
&
& \multicolumn{5}{c||}{\textbf{LIBERO (\%)}}
& \multicolumn{2}{c}{\textbf{RoboTwin (\%)}} \\
\cmidrule(lr){4-8}
\cmidrule(lr){9-10}
\textbf{Image source}
& \textbf{IDM action}
& \textbf{Source}
& \textbf{Spatial}
& \textbf{Object}
& \textbf{Goal}
& \textbf{Long}
& \textbf{Avg.}
& \textbf{Easy}
& \textbf{Hard} \\
\midrule

Dataset image
& \checkmark
& Replay
& \tb{71.6}
& \ul{74.0}
& \tb{53.0}
& \tb{29.2}
& \tb{57.0}
& \tb{65.3}
& \tb{60.7}  \\

\midrule

WAN2.2
& \checkmark
& Zero-shot
& 0.2
& 0.8
& 0.2
& 0.0
& 0.3
& 1.9
& 1.3  \\

Cosmos
& \checkmark
& Zero-shot
& 0.4
& 0.6
& 0.4
& 0.0
& 0.4
& 1.6
& 1.1  \\

\midrule

\rowcolor{blue!4}
WAN2.2 + VITA
& \checkmark
& Zero-shot
& \ul{64.4} {\scriptsize\textcolor{green!50!black}{(+64.2)}}
& 65.2 {\scriptsize\textcolor{green!50!black}{(+64.4)}}
& 36.8 {\scriptsize\textcolor{green!50!black}{(+36.6)}}
& 21.4 {\scriptsize\textcolor{green!50!black}{(+21.4)}}
& 47.0 {\scriptsize\textcolor{green!50!black}{(+46.7)}}
& \ul{59.6} {\scriptsize\textcolor{green!50!black}{(+57.7)}}
& \ul{40.4} {\scriptsize\textcolor{green!50!black}{(+39.1)}} \\

\rowcolor{blue!4}
Cosmos + VITA
& \checkmark
& Zero-shot
& 63.6 {\scriptsize\textcolor{green!50!black}{(+63.2)}}
& \tb{79.2} {\scriptsize\textcolor{green!50!black}{(+78.6)}}
& \ul{40.6} {\scriptsize\textcolor{green!50!black}{(+40.2)}}
& \ul{23.2} {\scriptsize\textcolor{green!50!black}{(+23.2)}}
& \ul{51.7} {\scriptsize\textcolor{green!50!black}{(+51.3)}}
& 57.7 {\scriptsize\textcolor{green!50!black}{(+56.1)}}
& 36.4 {\scriptsize\textcolor{green!50!black}{(+35.3)}} \\

\bottomrule
\end{tabular}
}
\end{table*}

\begin{table*}[t]
\centering
\caption{\textbf{Effect of VITA supervision under cross-environment
evaluation on LIBERO-Plus.}
All world models are trained on LIBERO and evaluated on LIBERO-Plus.
Zero-shot rows use future videos generated by the corresponding world
model, while the replay row uses recorded dataset observations.
Rows with ``+VITA'' use the additional action-grounded supervision.
The checkmark indicates that the frozen AoE model is used to recover
actions from images. \textcolor{green!50!black}{Green}/\textcolor{red}{Red} numbers show the \textcolor{green!50!black}{improvement}/\textcolor{red}{degradation} over baseline.}
\label{tab:libero_plus_vita}
\renewcommand{\arraystretch}{1.15}
\resizebox{\linewidth}{!}{
\newcolumntype{R}{>{\hspace{0em}}c<{\hspace{0.575em}}}
\begin{tabular}{l | c R || c c c c c c c | c}
\toprule
\textbf{Image source}
& \textbf{IDM action}
& \textbf{Source}
& \textbf{Background}
& \textbf{Light}
& \textbf{Language}
& \textbf{Noise}
& \textbf{Layout}
& \textbf{Camera}
& \textbf{Robot Init.}
& \textbf{Overall} \\
\midrule

Dataset image
& \checkmark
& Replay
& \tb{60.7}
& \tb{60.3}
& \tb{59.6}
& 0.0
& \tb{53.6}
& \tb{57.0}
& --
& \tb{47.0}  \\

\midrule

WAN2.2
& \checkmark
& Zero-shot
& 0.9
& 0.5
& 0.5
& 0.0
& 0.5
& 0.7
& 0.3
& 0.5  \\

Cosmos
& \checkmark
& Zero-shot
& 1.1
& 0.4
& 0.5
& 0.0
& 0.1
& 0.4
& 0.3
& 0.4  \\

\midrule

\rowcolor{blue!4}
WAN2.2 + VITA
& \checkmark
& Zero-shot
& 38.5 {\scriptsize\textcolor{green!50!black}{(+37.6)}}
& 45.3 {\scriptsize\textcolor{green!50!black}{(+44.8)}}
& \ul{37.1} {\scriptsize\textcolor{green!50!black}{(+36.6)}}
& \tb{0.1} {\scriptsize\textcolor{green!50!black}{(+0.1)}}
& 39.7 {\scriptsize\textcolor{green!50!black}{(+39.2)}}
& 26.6 {\scriptsize\textcolor{green!50!black}{(+25.9)}}
& \ul{13.4} {\scriptsize\textcolor{green!50!black}{(+13.1)}}
& 29.9 {\scriptsize\textcolor{green!50!black}{(+29.4)}} \\

\rowcolor{blue!4}
Cosmos + VITA
& \checkmark
& Zero-shot
& \ul{40.6} {\scriptsize\textcolor{green!50!black}{(+39.5)}}
& \ul{48.2} {\scriptsize\textcolor{green!50!black}{(+47.8)}}
& 34.1 {\scriptsize\textcolor{green!50!black}{(+33.6)}}
& 0.0 {\scriptsize\textcolor{green!50!black}{(+0.0)}}
& \ul{33.8} {\scriptsize\textcolor{green!50!black}{(+33.7)}}
& \ul{33.8} {\scriptsize\textcolor{green!50!black}{(+33.4)}}
& \textbf{28.6} {\scriptsize\textcolor{green!50!black}{(+28.3)}}
& \ul{30.3} {\scriptsize\textcolor{green!50!black}{(+29.9)}} \\

\bottomrule
\end{tabular}
}
\end{table*}

\begin{table*}[t]
\centering
\caption{\textbf{In-distribution performance of WAMs with VITA.}
Models are trained on LIBERO and RoboTwin clean+random $\rightarrow$ clean+random. \textcolor{green!50!black}{Green}/\textcolor{red}{Red} numbers show the \textcolor{green!50!black}{improvement}/\textcolor{red}{degradation} over baseline.
}
\label{tab:wam_vita_indomain}
\renewcommand{\arraystretch}{1.15}
\resizebox{0.85\linewidth}{!}{
\newcolumntype{R}{>{\hspace{0.975em}}c<{\hspace{0.975em}}}

\begin{tabular}{l | cccc | c || cc | c}
\toprule
& \multicolumn{5}{c||}{\textbf{LIBERO}}
& \multicolumn{3}{c}{
\textbf{RoboTwin: C+R $\rightarrow$ C+R}
} \\
\cmidrule(lr){2-6}
\cmidrule(lr){7-9}
\textbf{Method}
& \textbf{Long}
& \textbf{Goal}
& \textbf{Object}
& \textbf{Spatial}
& \textbf{Avg.}
& \textbf{Easy}
& \textbf{Hard}
& \textbf{Avg.} \\
\midrule

FastWAM
& 95.2
& 97.0
& \tb{100.0}
& 98.2
& 97.6
& 86.4
& 88.2
& 87.3 \\

LingBot-VA
& \ul{98.5}
& 97.2
& 99.6
& \ul{98.5}
& 98.5
& 85.6
& 84.2
& 84.9 \\
\midrule
\rowcolor{blue!4}
FastWAM + VITA
& 97.2 {\scriptsize\textcolor{green!50!black}{(+2.0)}}
& \tb{98.8} {\scriptsize\textcolor{green!50!black}{(+1.8)}}
& 99.6 {\scriptsize\textcolor{red}{(-0.4)}}
& \tb{99.0} {\scriptsize\textcolor{green!50!black}{(+0.8)}}
& \ul{98.7} {\scriptsize\textcolor{green!50!black}{(+1.1)}}
& \ul{90.7} {\scriptsize\textcolor{green!50!black}{(+4.3)}}
& \tb{91.9} {\scriptsize\textcolor{green!50!black}{(+3.7)}}
& \ul{91.3} {\scriptsize\textcolor{green!50!black}{(+4.0)}} \\

\rowcolor{blue!4}
LingBot-VA + VITA
& \tb{98.8} {\scriptsize\textcolor{green!50!black}{(+0.3)}}
& \ul{98.4} {\scriptsize\textcolor{green!50!black}{(+1.2)}}
& \ul{99.8} {\scriptsize\textcolor{green!50!black}{(+0.2)}}
& \tb{99.0} {\scriptsize\textcolor{green!50!black}{(+0.5)}}
& \tb{99.0} {\scriptsize\textcolor{green!50!black}{(+0.5)}}
& \tb{92.4} {\scriptsize\textcolor{green!50!black}{(+6.8)}}
& \ul{91.2} {\scriptsize\textcolor{green!50!black}{(+7.0)}}
& \tb{91.8} {\scriptsize\textcolor{green!50!black}{(+6.9)}} \\

\bottomrule
\end{tabular}
}
\end{table*}

\begin{table*}[t]
\centering
\caption{\textbf{Out-of-distribution performance of WAMs with VITA.}
Models are trained on LIBERO and RoboTwin clean environments, then
evaluated on LIBERO-Plus and RoboTwin clean+random environments,
respectively. Results are success rates in percent.
RoboTwin Avg. is the unweighted mean of Easy and Hard.
\textcolor{green!50!black}{Green}/\textcolor{red}{Red} numbers show the
\textcolor{green!50!black}{improvement}/\textcolor{red}{degradation}
over the corresponding baseline in percentage points.}
\label{tab:wam_vita_ood}
\renewcommand{\arraystretch}{1.15}
\resizebox{\linewidth}{!}{
\newcolumntype{R}{>{\hspace{0.975em}}c<{\hspace{0.975em}}}
\begin{tabular}{l | c c c c c c c | c || cc | c}
\toprule
& \multicolumn{8}{c||}{\textbf{LIBERO-Plus}}
& \multicolumn{3}{c}{\textbf{RoboTwin:
C $\rightarrow$ C+R}} \\
\cmidrule(lr){2-9}
\cmidrule(lr){10-12}
\textbf{Method}
& \textbf{Background}
& \textbf{Light}
& \textbf{Language}
& \textbf{Noise}
& \textbf{Layout}
& \textbf{Robot Init.}
& \textbf{Camera}
& \textbf{Overall}
& \textbf{Easy}
& \textbf{Hard}
& \textbf{Avg.} \\
\midrule

FastWAM
& 53.7
& 78.2
& 68.9
& \tb{37.7}
& \ul{60.7}
& 44.5
& \tb{16.4}
& \ul{51.5}
& 69.5
& 9.3
& 39.4 \\

LingBot-VA
& 56.3
& 46.5
& 61.7
& 11.4
& 34.8
& 36.2
& 9.5
& 35.1
& \ul{74.3}
& \ul{19.2}
& \ul{46.8} \\
\midrule
\rowcolor{blue!4}
FastWAM + VITA
& \ul{74.6} {\scriptsize\textcolor{green!50!black}{(+20.9)}}
& \tb{80.5} {\scriptsize\textcolor{green!50!black}{(+2.3)}}
& \tb{75.6} {\scriptsize\textcolor{green!50!black}{(+6.7)}}
& \ul{27.4} {\scriptsize\textcolor{red}{(-10.3)}}
& \tb{80.9} {\scriptsize\textcolor{green!50!black}{(+20.2)}}
& \tb{53.4} {\scriptsize\textcolor{green!50!black}{(+8.9)}}
& \ul{14.2} {\scriptsize\textcolor{red}{(-2.2)}}
& \tb{55.9} {\scriptsize\textcolor{green!50!black}{(+4.4)}}
& 73.8 {\scriptsize\textcolor{green!50!black}{(+4.3)}}
& 17.5 {\scriptsize\textcolor{green!50!black}{(+8.2)}}
& 45.7 {\scriptsize\textcolor{green!50!black}{(+6.3)}} \\

\rowcolor{blue!4}
LingBot-VA + VITA
& \tb{77.9} {\scriptsize\textcolor{green!50!black}{(+21.6)}}
& \ul{71.2} {\scriptsize\textcolor{green!50!black}{(+24.7)}}
& \ul{69.9} {\scriptsize\textcolor{green!50!black}{(+8.2)}}
& 15.0 {\scriptsize\textcolor{green!50!black}{(+3.6)}}
& 51.2 {\scriptsize\textcolor{green!50!black}{(+16.4)}}
& \ul{49.3} {\scriptsize\textcolor{green!50!black}{(+13.1)}}
& 12.7 {\scriptsize\textcolor{green!50!black}{(+3.2)}}
& 47.2 {\scriptsize\textcolor{green!50!black}{(+12.1)}}
& \tb{82.5} {\scriptsize\textcolor{green!50!black}{(+8.2)}}
& \tb{31.6} {\scriptsize\textcolor{green!50!black}{(+12.4)}}
& \tb{57.1} {\scriptsize\textcolor{green!50!black}{(+10.3)}} \\

\bottomrule
\end{tabular}
}
\end{table*}

\subsection{VITA Supervision Makes Futures Executable}
\label{sec:vita_results}

\paragraph{Cross-task executability.}
We add the VITA loss of Section~\ref{sec:video_action_loss} to concentrate generation on action-relevant regions. Table~\ref{tab:vita_future_video_execution} shows that this simple change makes generated futures executable: WAN2.2
reaches 47.0\% average success on LIBERO and Cosmos 51.7\%, approaching the
57.0\% of recorded images, and RoboTwin Hard rises from around 1\% to 40.4\%
and 36.4\%, respectively. The gains persist across environments
(Table~\ref{tab:libero_plus_vita}), where both models reach about 30\%
overall on LIBERO-Plus. To test whether this improvement depends on using AoE as both the VITA teacher and the execution evaluator, we keep the same VITA-supervised WAN2.2 checkpoint and replace only the IDM used after video generation. We further evaluate VITA-trained futures with J-IDM~\citep{vera} and the DreamGen IDM~\citep{dreamgen}; detailed results are reported in Appendix~\ref{app:teacher_independence}.

\paragraph{VITA for world action models.}
Adding $\mathcal{L}_{\mathrm{VITA}}$ to LingBot-VA and FastWAM training
improves both architectures in-distribution
(Table~\ref{tab:wam_vita_indomain}); on RoboTwin, for example, LingBot-VA
rises from 84.9\% to 91.8\% average success. The benefits are substantially
larger out of distribution (Table~\ref{tab:wam_vita_ood}): LingBot-VA
improves from 35.1\% to 47.2\% overall on LIBERO-Plus, with the largest
gains under background and lighting changes, and raises RoboTwin Hard success from 19.2\% to 31.6\%. FastWAM follows the same trend, which is notable
because its inference path never renders future images, and the improvement
arrives purely through training, as analyzed in
Section~\ref{sec:supervision_analysis}.


\begin{figure*}[t]
  \centering
  \includegraphics[width=\textwidth]{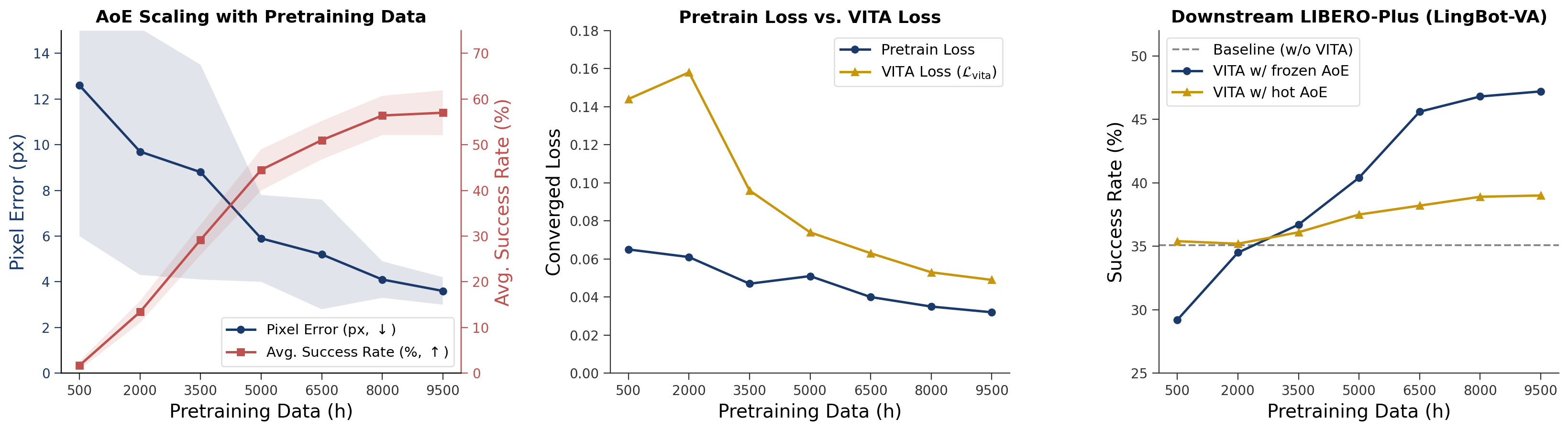}
  \caption{\textbf{AoE pretraining scaling: effects on cross-task action recovery and VITA-based policy learning.}
  \emph{Left:} IDM's gripper pixel error and recovered-action success on cross-task dataset image replay task as pretraining data grows.
  \emph{Middle:} WM's VITA loss across data
  scales with the converged pretrain loss.
  \emph{Right:} WAM's downstream LIBERO-Plus success with
  the VITA loss, with AoE frozen or updated (hot); the dashed line marks WAM without VITA.}
  \label{fig:aoe_scaling}
\end{figure*}

\subsection{Scaling AoE Pretraining}
\label{sec:aoe_scaling}

Both the evaluation and the supervision above rely on AoE, so we study how
its pretraining data scale affects each role by partitioning the corpus into
mini-sets and adding them successively, spanning 500 to 9{,}500 hours. As
Figure~\ref{fig:aoe_scaling} (left) shows, the gripper pixel error on
held-out tasks drops from 12.6\,px to 3.6\,px while the zero-shot success rises from 1.7\% to 57.0\%.
The converged VITA loss descends together with the pretrain loss across data
scales (middle), remaining a well-behaved proxy as the teacher improves.
Downstream (right), LingBot-VA's LIBERO-Plus success with a frozen AoE grows
monotonically with teacher quality and overtakes the no-VITA baseline
between 3{,}500 and 5{,}000 hours, whereas a trainable AoE
stays nearly flat. Since updating the teacher erodes the calibrated interface
that makes the supervision meaningful, AoE remains frozen in all
experiments.


\subsection{Ablations}
\label{sec:ablations}

We ablate three design choices: the camera-space action representation, the
matching component of the VITA loss, and the cost of camera-space actions
when cross-task transfer is not required.

\begin{table*}[t]
\centering
\renewcommand{\arraystretch}{1.15}
\begin{minipage}[t]{0.56\linewidth}
\centering
\caption{\textbf{Camera-space vs.\ 3D-pose action representations for inverse dynamics.} Pretrain loss and validation RMSE measure action-prediction quality ($\downarrow$), while cross-task success rates measure downstream execution ($\uparrow$). }
\label{tab:idm_action_representation}
\resizebox{\linewidth}{!}{
\begin{tabular}{l | c c | c c}
\toprule
& \multicolumn{2}{c|}{\textbf{Pretraining metrics}}
& \multicolumn{2}{c}{\textbf{Cross-task SR (\%) $\uparrow$}} \\
\cmidrule(lr){2-3}
\cmidrule(lr){4-5}
\textbf{Action repr.}
& \textbf{Pretrain loss $\downarrow$}
& \textbf{Valid RMSE $\downarrow$}
& \textbf{Zero-shot}
& \textbf{10-shot} \\
\midrule

\rowcolor{blue!4}
Camera-space (AoE)
& 0.047
& \tb{0.070}
& \tb{29.2}
& \tb{37.8} \\

3D pose
& \tb{0.026}
& 0.146
& 0.0
& 2.3 \\

\bottomrule
\end{tabular}
}
\end{minipage}
\hfill
\begin{minipage}[t]{0.41\linewidth}
\centering
\caption{\textbf{Ablation of the VITA loss components on LingBot-VA.}
Success rates on LIBERO-Plus and the hard split of RoboTwin when training
with each aspect of VITA
loss.}
\label{tab:vita_loss_ablation}
\resizebox{\linewidth}{!}{
\begin{tabular}{l | c c}
\toprule
\textbf{Model}
& \textbf{LIBERO-Plus (\%)}
& \textbf{RoboTwin Hard (\%)} \\
\midrule

LingBot-VA
& 35.1
& 19.2 \\

\quad + matching loss only
& 36.7
& 23.9 \\

\quad + GT loss only
& \ul{45.5}
& \ul{24.7} \\

\rowcolor{blue!4}
\quad + VITA loss (Ours)
& \tb{47.2}
& \tb{31.6} \\

\bottomrule
\end{tabular}
}
\end{minipage}
\end{table*}

\paragraph{Camera-space vs.\ 3D-pose actions.}
The 3D-pose variant fits its training environments better than AoE, but this
fit comes from memorizing environment-specific coordinate conventions: it
recovers executable actions on unseen tasks at 0.0\% zero-shot and 2.3\%
with 10-shot adaptation, where the camera-space representation reaches
29.2\% and 37.8\% (Table~\ref{tab:idm_action_representation}). Because the camera-space target is defined by observable image geometry, the same action interface transfers across environments without relearning a task-specific world-frame mapping.
\begin{wrapfigure}{r}{0.37\textwidth}
  \centering
  \vspace{4pt}
  \includegraphics[width=0.37\textwidth]{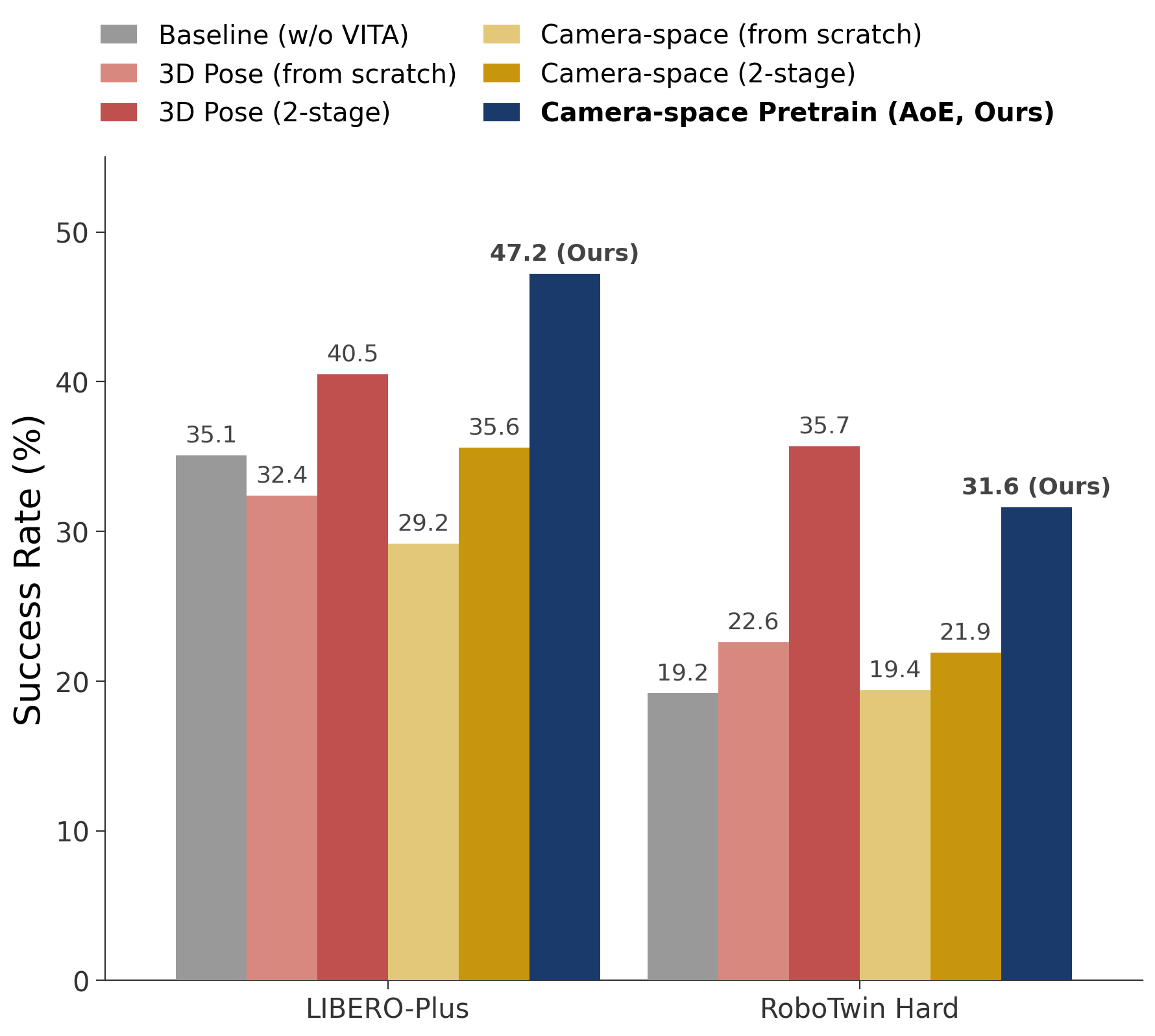}
  \caption{\textbf{VITA loss computed with different IDM variants.} IDM is not suitable for updating with world models. 3D pose performs better, especially in environments with severe screen occlusion.}
  \label{fig:idm_ablation}
  \vspace{-20pt}

\end{wrapfigure}
\paragraph{What does camera space cost without transfer?}
Figure~\ref{fig:idm_ablation} compares six IDM variants as VITA supervisors within a fixed environment. Pretraining matters more than the representation: from-scratch IDMs of either type supervise poorly, while a two-stage trained 3D-pose IDM is competitive. Weak supervision is worse than none: both from-scratch variants fall on LIBERO-Plus below the baseline trained without VITA, indicating that an uncalibrated teacher actively misdirects the world model.
Adapting the IDM to the downstream environment in a second stage recovers much of this deficit, yet still lags large-scale pretraining on LIBERO-Plus, suggesting that per-environment adaptation alone does not match the supervision quality of a broadly pretrained teacher. 3D pose performs better because camera-space supervision becomes unreliable under severe occlusion. The camera-space pretrained AoE remains the best supervisor on LIBERO-Plus at 47.2\%, and trails 3D-pose 2-stage only on RoboTwin Hard. 
\paragraph{Why the matching loss.}
The GT loss alone reaches 45.5\% on LIBERO-Plus but only 24.7\% on RoboTwin Hard (Table~\ref{tab:vita_loss_ablation}). The reason is structural: when the gripper leaves the camera view or the camera undergoes deformation, camera-space GT targets become undefined and provide little signal. The matching loss transfers action-relevant information from the recorded future without requiring a pose estimate and further disambiguates multi-solution observations, so the full VITA objective performs best on both benchmarks.

\section{Related Work}
\label{sec:related}
\paragraph{Grounded world-model evaluation.}
A growing body of work evaluates world models in embodied scenarios. WorldArena jointly scores perceptual quality and functional utility, revealing a persistent gap between visual fidelity and embodied usefulness~\citep{worldarena}. KineBench grounds generated videos kinematically, extracting end-effector poses from individual frames for closed-loop execution and trajectory-level analysis~\citep{kinebench}, and WorldSimProbe issues controlled simulator queries to test action realization and interaction grounding~\citep{worldsimprobe}. Despite this progress, existing evaluations remain confined to imagination: they score the generated videos themselves, or execute them only through extractors tied to their training environments, leaving unquantified whether a predicted future can be carried into an unseen environment and actually executed~\citep{jiang2026robowmbenchbenchmarkevaluatingworld}. We therefore use dataset-image and recorded-action replay to calibrate IDM transfer, and hold the pretrained IDM fixed across visual-source replacements, human judgments, and local geometric interventions.

\paragraph{Video and inverse-dynamics priors for control.}
DreamGen converts generated videos into neural trajectories for policy training~\citep{dreamgen}. LingBot-VA couples causal video modeling with action prediction~\citep{lingbotva}, while Fast-WAM studies video co-training under several inference interfaces~\citep{fastwam}. VERA combines video planning with a Jacobian-based inverse dynamics model~\citep{vera}. GE-Act 2.0 separately pretrains a visual planner and IDM, then jointly optimizes them through generated future states using action flow matching and compatibility-based selection~\citep{geact2}. These approaches establish visual prediction and inverse dynamics as complementary components of robot learning. We evaluate a pretrained, frozen camera-space IDM through dataset-image replay on held-out tasks and transfer its action knowledge into future-image supervision.

\section{Conclusion}
\label{sec:conclusion}

Pretrained inverse dynamics provides a common action model for evaluating and supervising visual futures. We introduce AoE, an inverse-dynamics model built on a camera-projection action representation with zero-shot cross-task generalization, through which world-model futures are carried from video into interactive execution in the environment. Building on this interface, we further propose the VITA loss, which substantially improves the cross-task generalization of world models. We also extend VITA to world action models, obtaining consistent gains across multiple architectures and evaluation environments. Constrained by the camera-projection representation, AoE cannot handle tasks where the gripper leaves the camera view for extended periods or the camera is severely occluded, and would raise additional safety concerns in real-robot evaluation; we expect these limitations to be addressed in combination with future work on active exploration.
\subsection*{AI use statement}
In this work, we used generative AI tools to assist with checking the training and evaluation data pipelines and code, polishing the language of the manuscript, and developing and checking the mathematical derivations for the VITA loss. We have not used generative AI tools for any other tasks requiring disclosure, and the remaining required disclosure categories are not applicable to this work. All AI-assisted work was reviewed and confirmed by the main authors against the underlying code, data, experiments, and mathematical derivations. We take full responsibility for the final content of this work, including all text, claims, code, and artifacts produced with the aid of generative AI.
\subsection*{Reproducibility Statement}
We provide detailed training and evaluation configurations in
Appendix~\ref{app:training_details}, including the composition and
proportions of the pretraining data in
Appendix~\ref{app:canonical_action_pretraining}, and detailed experimental
results in Appendix~\ref{app:full_results}. The equations for converting
camera-space predictions into 3D poses in the robot coordinate frame
are provided in Appendix~\ref{app:camera_action}. A formal analysis of
the VITA loss, including its assumptions and proofs, is presented in
Appendix~\ref{app:supervision_analysis}. The supplementary materials
include all training and evaluation code, together with the code for
converting and cleaning the pretraining data. We also plan to publicly
release the pretrained AoE checkpoint in the future.
\bibliography{iclr2027_conference}
\bibliographystyle{iclr2027_conference}
\appendix

\section{Attention Maps and Camera-Space Representations on Seen and Unseen Tasks}
\label{app:visualization}

We visualize AoE on two types of data: a validation split sampled from
the pretraining dataset and tasks from unseen environments. The same
frozen AoE checkpoint is used in both cases.

\begin{figure}[t]
    \centering
    \includegraphics[width=\linewidth]{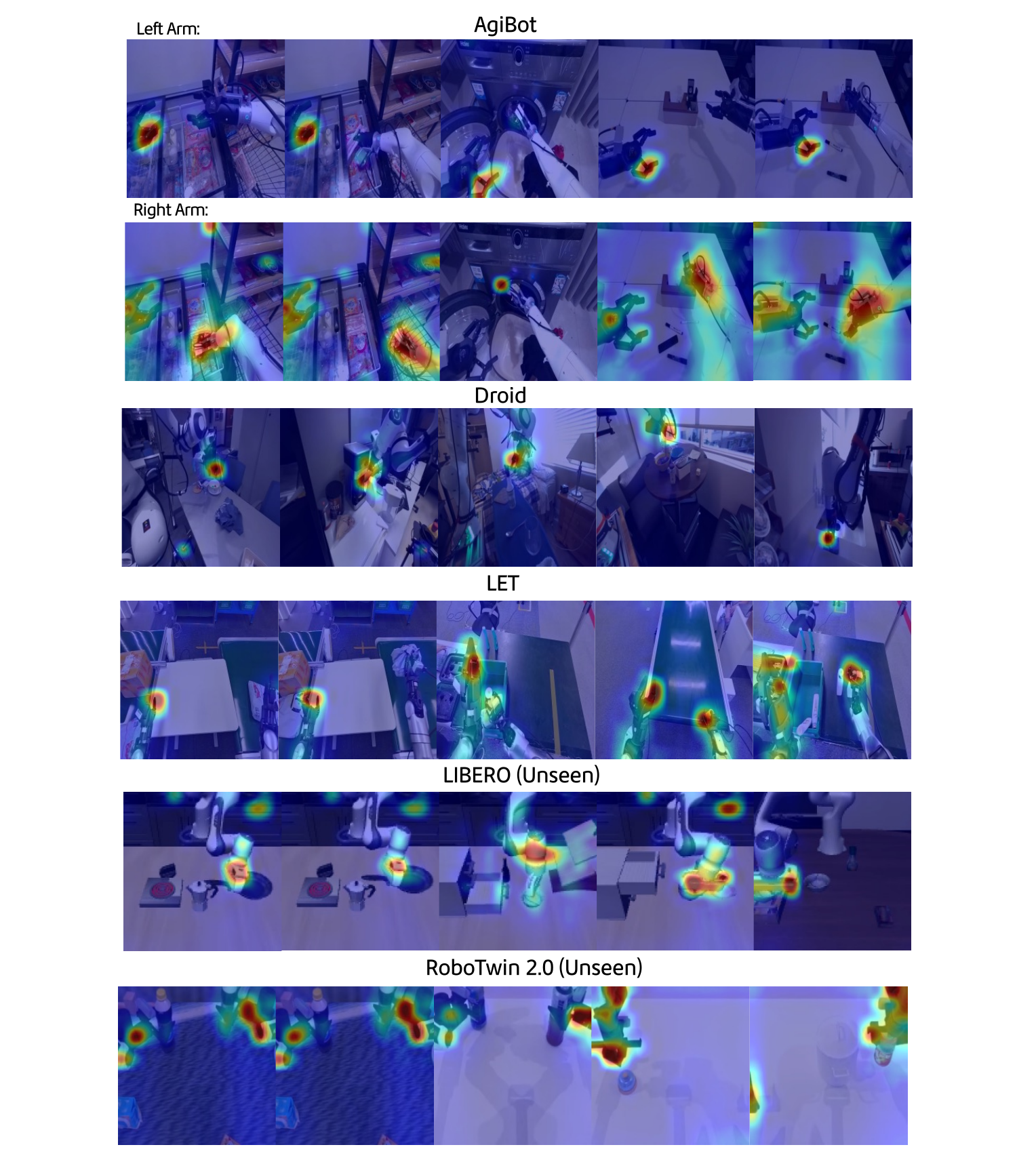}
    \caption{\textbf{Attention maps on seen and unseen tasks.}
    AoE focuses primarily on the robot end effector and the object being
    manipulated in both the validation split and unseen environments.}
    \label{fig:attn}
\end{figure}

\paragraph{Attention maps.}
Figure~\ref{fig:attn} shows attention maps for the validation split and
unseen tasks. Across both settings, the attention is concentrated around
the robot end effector and the object being manipulated, while receiving
less emphasis from background regions. This behavior is consistent with
the action-grounded sensitivity analyzed in Section~\ref{sec:supervision_analysis}:
visual changes near the end effector and the target object have a direct
effect on the action-flow prediction. The attention maps provide a
qualitative visualization of this behavior, while the formal analysis
in Section~\ref{sec:supervision_analysis} is based on the local
sensitivity of AoE's action-flow output to its visual input.

\begin{figure}[t]
    \centering
    \includegraphics[width=\linewidth,keepaspectratio]{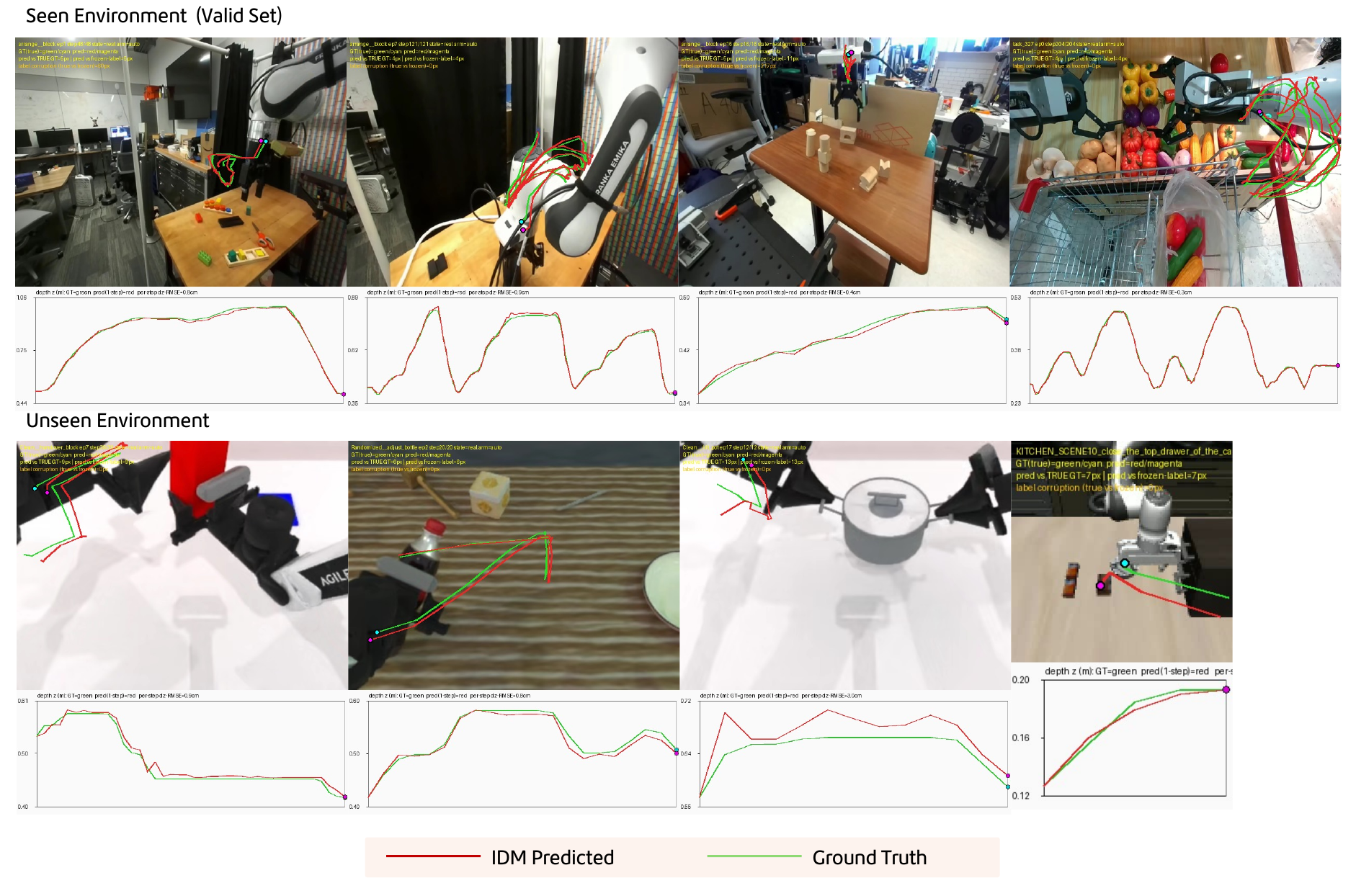}
    \caption{\textbf{Camera-space representation on seen and unseen
    tasks.}
    The overlaid trajectories are generated by AoE from recorded video
    pairs, and the plots below show the corresponding predicted
    relative-depth changes.}
    \label{fig:traj}
\end{figure}

\paragraph{Camera-space representations.}
Figure~\ref{fig:traj} visualizes the camera-space action representation
predicted by AoE. The examples are obtained from recorded videos in the
validation split and unseen environments. For each observation pair,
AoE predicts the two-dimensional end-effector trajectory in the camera
image, shown as the overlaid line between the recorded start and end
frames. The plot below each example shows the corresponding predicted
relative-depth change. These visualizations show that AoE produces a
consistent camera-space representation across seen and unseen tasks,
without requiring the action target to be expressed in a
task-specific world coordinate frame.

\section{Camera-Space Action Representation and Pose Reconstruction}
\label{app:camera_action}

This appendix specifies the camera-space action targets used to pretrain AoE and their conversion into executable pose commands. We use subscripts $s$ and $e$ for the start and end of an observation pair. Superscripts $c$ and $w$ denote the camera frame and the world or robot-base frame used by the controller, respectively.

\paragraph{Endpoint representation.}

For each arm $r \in \mathcal{R}$, let $\mathbf{p}^{r}_{s}$ and $\mathbf{p}^{r}_{e}$ denote the start and end positions of the end effector. We summarize the translational component of the transition by the segment
\begin{equation}
    \mathbf{p}^{r}(\alpha)
    =
    (1-\alpha)\mathbf{p}^{r}_{s}
    +
    \alpha\mathbf{p}^{r}_{e},
    \qquad \alpha \in [0,1].
    \label{eq:app_endpoint_segment}
\end{equation}
The action target also retains the relative orientation change and the endpoint gripper command. The segment describes the geometric displacement represented by the target; intermediate waypoints and their timing are omitted from this representation.

\paragraph{Projection into the camera frame.}

Let $K$ denote the camera intrinsic matrix, and let $(R_{cw},\mathbf{t}_{cw})$ transform positions from the controller frame into the camera frame. For endpoint $j \in \{s,e\}$,
\begin{equation}
    \mathbf{p}^{c,r}_{j}
    =
    R_{cw}\mathbf{p}^{w,r}_{j}
    +
    \mathbf{t}_{cw},
    \qquad
    \begin{bmatrix}
        u^{r}_{j} \\ v^{r}_{j} \\ 1
    \end{bmatrix}
    =
    \frac{1}{z^{r}_{j}}K\mathbf{p}^{c,r}_{j},
    \label{eq:app_camera_projection}
\end{equation}
where $z^{r}_{j}$ is the third coordinate of $\mathbf{p}^{c,r}_{j}$. The intrinsic matrix is expressed in the same pixel coordinate system as the input images.

For an image of width $W$ and height $H$, we normalize the projected coordinates as
\begin{equation}
    \bar{u}^{r}_{j} = \frac{2u^{r}_{j}}{W}-1,
    \qquad
    \bar{v}^{r}_{j} = \frac{2v^{r}_{j}}{H}-1.
    \label{eq:app_pixel_normalization}
\end{equation}
The depth target records the change along the camera's depth axis:
\begin{equation}
    \Delta z^{r} = z^{r}_{e}-z^{r}_{s},
    \qquad
    \widetilde{\Delta z}^{\,r}
    =
    \mathcal{N}_{z}(\Delta z^{r}),
    \label{eq:app_depth_target}
\end{equation}
where $\mathcal{N}_{z}$ denotes the depth normalization used during pretraining. The same normalization and its inverse are used at evaluation.

\paragraph{Rotation and gripper targets.}

Let $\mathbf{R}^{c,r}_{s}$ and $\mathbf{R}^{c,r}_{e}$ denote the end-effector orientations in the camera frame. We write the relative-rotation target as
\begin{equation}
    \boldsymbol{\rho}^{r}
    =
    \operatorname{Enc}_{\mathrm{rot}}
    \left(
        \operatorname{Rel}
        (\mathbf{R}^{c,r}_{s},\mathbf{R}^{c,r}_{e})
    \right).
    \label{eq:app_rotation_target}
\end{equation} The 6D rotation encoding follows \citet{rotation6d}.
Here, $\operatorname{Rel}$ follows the relative-rotation convention used to construct the training labels, and $\operatorname{Enc}_{\mathrm{rot}}$ maps the resulting rotation to its continuous representation. We use a six-dimensional rotation encoding. Its decoder, $\operatorname{Dec}_{\mathrm{rot}}$, is paired with the same rotation convention during pose reconstruction. The scalar $g^{r}$ denotes the endpoint gripper command.

The complete target for one arm is
\begin{equation}
    \tilde{\mathbf{a}}^{r}
    =
    \left[
        \bar{u}^{r}_{s},
        \bar{v}^{r}_{s},
        \bar{u}^{r}_{e},
        \bar{v}^{r}_{e},
        \widetilde{\Delta z}^{\,r},
        \boldsymbol{\rho}^{r},
        g^{r}
    \right]
    \in \mathbb{R}^{d_a}.
    \label{eq:app_complete_action}
\end{equation}
In our implementation, $d_a=12$: four projected endpoint coordinates, one depth-change value, six rotation coordinates, and one gripper value. Each arm has its own action target.

\paragraph{Reconstructing the endpoint position.}

The current robot state $s_t$ supplies the initial end-effector position and orientation through forward kinematics. The initial camera-frame depth is
\begin{equation}
    z^{r}_{s}
    =
    \left[
        R_{cw}\mathbf{p}^{w,r}_{s}
        +
        \mathbf{t}_{cw}
    \right]_{z}.
    \label{eq:app_initial_depth}
\end{equation}
Given the predicted normalized endpoint coordinates and depth change, we recover
\begin{equation}
    \hat{u}^{r}_{e}
    =
    \frac{W}{2}(\hat{\bar{u}}^{r}_{e}+1),
    \qquad
    \hat{v}^{r}_{e}
    =
    \frac{H}{2}(\hat{\bar{v}}^{r}_{e}+1),
    \qquad
    \hat{z}^{r}_{e}
    =
    z^{r}_{s}
    +
    \mathcal{N}_{z}^{-1}
    \left(\widehat{\widetilde{\Delta z}}^{\,r}\right).
    \label{eq:app_endpoint_denormalization}
\end{equation}
Back-projection gives the endpoint in the camera frame, followed by conversion into the controller frame:
\begin{equation}
    \hat{\mathbf{p}}^{c,r}_{e}
    =
    \hat{z}^{r}_{e}K^{-1}
    \begin{bmatrix}
        \hat{u}^{r}_{e} \\
        \hat{v}^{r}_{e} \\
        1
    \end{bmatrix},
    \qquad
    \hat{\mathbf{p}}^{w,r}_{e}
    =
    R_{cw}^{\top}
    \left(
        \hat{\mathbf{p}}^{c,r}_{e}
        -
        \mathbf{t}_{cw}
    \right).
    \label{eq:app_endpoint_reconstruction}
\end{equation}

\paragraph{Reconstructing the endpoint orientation.}

The initial orientation and predicted relative rotation determine the endpoint orientation:
\begin{equation}
    \hat{\mathbf{R}}^{c,r}_{e}
    =
    \operatorname{Apply}
    \left(
        \mathbf{R}^{c,r}_{s},
        \operatorname{Dec}_{\mathrm{rot}}
        (\hat{\boldsymbol{\rho}}^{r})
    \right),
    \qquad
    \hat{\mathbf{R}}^{w,r}_{e}
    =
    R_{cw}^{\top}\hat{\mathbf{R}}^{c,r}_{e}.
    \label{eq:app_orientation_reconstruction}
\end{equation}
The operators $\operatorname{Rel}$ and $\operatorname{Apply}$ use matching conventions, satisfying
$\operatorname{Apply}(R_s,\operatorname{Rel}(R_s,R_e))=R_e$.
The reconstructed position, orientation, and gripper command are then passed to the environment's controller using its action convention.

\paragraph{Execution from video.}

For a generated video, AoE processes consecutive frame pairs. At each execution step, the current robot state supplies the initial pose required for reconstruction. Camera calibration, action normalization, and controller conventions remain fixed when comparing recorded and generated videos. All coordinate conversions are deterministic, and AoE retains its pretrained parameters throughout evaluation.

\section{Inverse-dynamics architecture and protocol details}
\label{app:probe_details}

\begin{figure}[t]
\centering
\includegraphics[width=\linewidth]{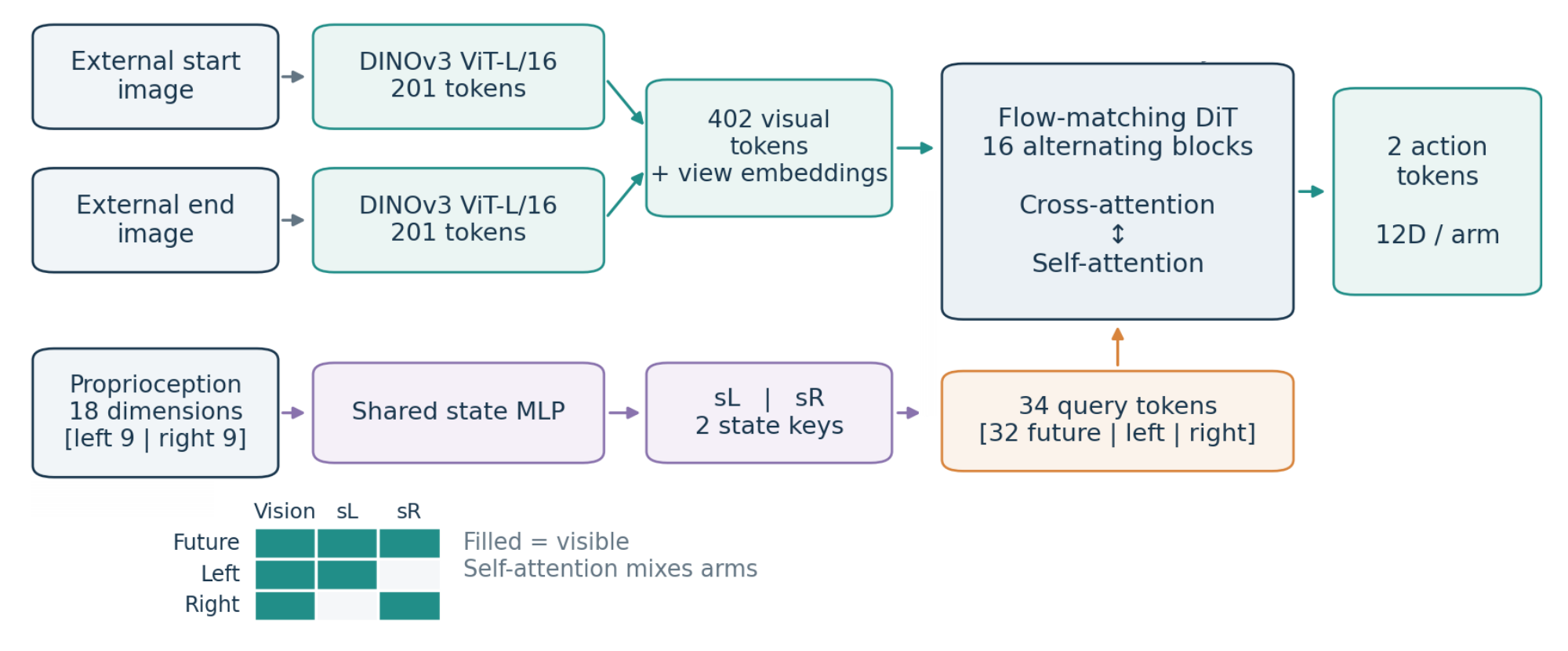}
\caption{\textbf{The AoE inverse-dynamics architecture.} Two independent image towers encode external-camera observations at the start and end of a transition. Alternating attention blocks connect shared visual features to two arm-action tokens. Direct state access is routed by arm; self-attention supports coordination. The output parameterizes translation in image coordinates and depth, with rotation and gripper control.}
\label{fig:architecture}
\end{figure}

\paragraph{Visual and action streams.}
AoE receives two external-camera images, \texttt{ext\_start} and \texttt{ext\_end}. Each timestamp has an independent DINOv3~\citep{dinov3} ViT-L/16 encoder; both full encoder parameter sets are updated during AoE pretraining. Each encoder produces 201 tokens: one class token, four register tokens, and 196 patch tokens on a $14\times14$ grid. Learned view embeddings distinguish the timestamps, producing 402 visual conditioning tokens. The model uses no wrist-camera input (Figure~\ref{fig:architecture}). 

The 18-dimensional proprioceptive vector is split into two nine-dimensional arm states. A shared state MLP produces tokens $s_L$ and $s_R$, bringing the conditioning stream to 404 tokens. The query stream has 34 tokens: 32 future-representation tokens and two 12-dimensional arm-action tokens projected into the transformer width. The action-token axis indexes arms. The transformer has width 1536, 16 layers, 32 attention heads of dimension 48, and MLP hidden width 1024, with approximately 450 million parameters. Together with both visual encoders, AoE contains approximately 1.14 billion parameters.

\paragraph{State routing and positional information.}
Even-indexed layers use cross-attention over the visual and state conditioning stream. The left action token attends to all visual tokens and $s_L$; the right action token attends to all visual tokens and $s_R$. The 32 future tokens attend to the complete conditioning stream. Odd-indexed layers apply unmasked self-attention across queries, allowing information exchange and bimanual coordination. Thus, the cross-attention mask routes direct state access, while self-attention permits indirect cross-arm influence.

Cross-attention uses two-dimensional axial rotary positional embeddings. Image patches carry their row and column coordinates, and each query token has a learned two-dimensional reference point. Both state tokens use coordinate $(0,0)$. The arm-supervision mask expands across all 12 action coordinates. If a minibatch contains multiple temporal pairs $k$, the numerator of the masked pretraining loss sums over $(b,k,r)$ and coordinates, and its denominator is $12\max(1,\sum_{b,k,r}m_{bkr})$. The same reduction applies at every sampled flow time. Unsupervised arms receive no direct loss contribution.

\subsection{Task splits and frozen-model reuse}
\label{app:task_splits}
LIBERO source tasks are selected from LIBERO-90 after excluding evaluation-task identities. The exclusion is applied before AoE pretraining. Dataset image replay evaluates the pretrained checkpoint on recorded transitions from these unseen tasks. The world-model source comparison reuses the same checkpoint for recorded, cross-task generated, and VITA-supervised generated futures. The VITA condition updates the world model alone.

RoboTwin uses two-fold task-level cross-validation with 25 training tasks and 25 held-out tasks per fold. The two groups exchange roles between folds. Each fold reuses its pretrained AoE checkpoint with frozen parameters. The complete task partition is given in Appendix~\ref{app:robotwin_task_partition}.

The native WAM transfer comparison trains on LIBERO-90 and evaluates the LIBERO Long, Goal, Object, and Spatial suites, with 500 episodes per suite and 2,000 episodes in total. The All training objectives in Table 20 share this split, including the native and DINO controls. The same-domain world-model comparison uses 500 episodes, with target adaptation supplied 50 demonstrations per task. Standard LIBERO and LIBERO-Plus native-control experiments use training on benchmark tasks and are reported as a separate evaluation setting. 

AoE parameters remain frozen after pretraining, except in the explicitly identified trainable-teacher scaling ablation in Section~\ref{sec:aoe_scaling}. Source-validation data provide model selection and the offline confidence weights used by the auxiliary loss. Confidence calibration changes loss weights while preserving the pretrained model. In downstream WAM training, differentiation through AoE's generated-image input transfers supervision to video and shared parameters. Each WAM uses its native action path for evaluation. The fold teacher and its confidence calibration are shared across downstream architecture comparisons.

\subsection{Calibration protocol}
\label{app:protocol_design}

Held-out task calibration fixes task identities before evaluation and excludes their demonstrations from AoE pretraining. Recorded actions and future images share the simulator snapshot and temporal interval. Camera perturbations are applied when rendering both images of a pair. We report full-set performance, the coverage and performance of instances with successful recorded-action replay, and their subset with successful demonstration-image recovery. An independent geometric recovery pipeline repeats the source comparison and the paired visual interventions. Rotation targets encode the start-to-end relative rotation in a consistent reference frame, and depth normalization is inverted before Eq.~\eqref{eq:app_endpoint_reconstruction}. Appendix~\ref{app:loss_details} specifies the complete auxiliary training objective, confidence calibration, and gradient schedule.

\section{Implementation of the vision-to-action Loss}
\label{app:video_action_loss}

\paragraph{Action-flow supervision.}

AoE is pretrained with flow matching~\citep{flowmatching}. Its supervised output is an action-update vector: the direction in which a noisy action estimate should move toward the target action. For a training pair $(o_t,o_{t+i})$, let $\tilde{\mathbf{a}}_{t,t+i}$ denote the ground-truth camera-space action defined in Appendix~\ref{app:camera_action}. We sample Gaussian noise $\boldsymbol{\epsilon}$ and a flow time $\tau$, and construct
\begin{equation}
    \mathbf{z}_{\tau}
    =
    (1-\tau)\boldsymbol{\epsilon}
    +
    \tau\tilde{\mathbf{a}}_{t,t+i},
    \qquad
    \mathbf{v}^{\star}_{t,i}
    =
    \tilde{\mathbf{a}}_{t,t+i}
    -
    \boldsymbol{\epsilon},
    \label{eq:app_action_flow}
\end{equation}
where $\boldsymbol{\epsilon}\sim\mathcal{N}(0,I)$ and
$\tau\sim\mathcal{U}[0,1]$. The vector $\mathbf{z}_{\tau}$ is the noisy action input, and $\mathbf{v}^{\star}_{t,i}$ is the corresponding target update.

The notation $f_{\phi}(o_t,o_{t+i})$ in the main text abbreviates AoE's flow prediction conditioned on the image pair, current robot state $s_t$, noisy action $\mathbf{z}_{\tau}$, and flow time $\tau$. The generated-image and recorded-image branches share all conditions except the future image. Their difference therefore measures how replacing the future image changes AoE's action prediction. The recorded-image branch is evaluated without gradient tracking.

\paragraph{Masked and confidence-weighted errors.}

The implementation uses weighted robust errors for the two objectives. For prediction $\mathbf{x}$, target $\mathbf{y}$, and nonnegative coordinate weights $\mathbf{w}$, we compute
\begin{equation}
    \mathcal{R}(\mathbf{x},\mathbf{y};\mathbf{w})
    =
    \frac{
        \sum_j w_j
        \left(
            \sqrt{1+(x_j-y_j)^2}-1
        \right)
    }{
        \sum_j w_j
    },
    \label{eq:app_robust_error}
\end{equation}
over samples with nonzero total weight. Samples without valid supervised coordinates are omitted.

The matching weights combine supervision validity, coordinate selection, and offline confidence calibrated on recorded transitions. The GT weights additionally account for the teacher's reliability on the corresponding recorded image pair. Coordinates on which the teacher has a larger error against the ground-truth flow target receive less weight in the GT objective. These weights are held fixed when differentiating with respect to generated images. The projected start coordinates remain part of AoE's action representation and receive zero direct auxiliary weight.

Using these weights, the implemented action-flow objectives are
\begin{equation}
\begin{aligned}
    \mathcal{L}^{\mathrm{rob}}_{\mathrm{match}}
    &=
    \mathbb{E}
    \left[
        \mathcal{R}
        \left(
            f_{\phi}(o_t,\hat{o}_{t+i}),
            f_{\phi}(o_t,o_{t+i});
            \mathbf{w}_{\mathrm{match}}
        \right)
    \right],
    \\
    \mathcal{L}^{\mathrm{rob}}_{\mathrm{GT}}
    &=
    \mathbb{E}
    \left[
        \mathcal{R}
        \left(
            f_{\phi}(o_t,\hat{o}_{t+i}),
            \mathbf{v}^{\star}_{t,i};
            \mathbf{w}_{\mathrm{GT}}
        \right)
    \right].
\end{aligned}
\label{eq:app_robust_action_losses}
\end{equation}

The GT objective can produce noisy image gradients when the frozen teacher is unreliable on generated observations. To provide a direct image-space anchor, we supplement the matching objective with a weak reconstruction term around the ground-truth gripper locations. This ROI term uses Gaussian spatial weights and a Charbonnier reconstruction error. It is included as an implementation-level stabilizer alongside the matching loss.

\paragraph{Gradient propagation and training.}

Generated future frames are decoded differentiably before being passed to AoE. AoE and the video decoder retain frozen parameters, while gradients through their inputs update the world model. The two auxiliary objectives are averaged over the supervised future frames and added to the native training objective. For WAMs, the native action objective is retained, and the auxiliary gradients reach the video branch and shared parameters. Evaluation follows the original action inference path.

\section{Analysis of vision-to-action Supervision}
\label{app:supervision_analysis}

\paragraph{Notation and scope.}
We first consider a single transition, holding the current observation, instruction, robot state, and world-model sampling inputs fixed. Write $y=o_{t+i}$ for the recorded future and $\hat y$ for the predicted future. Let $\zeta$ collect the shared AoE noising conditions, including the noisy action and flow time. We abbreviate the frozen teacher's action-flow prediction as
\begin{equation}
    F_\zeta(y)=f_\phi(o_t,y;\zeta),
    \qquad
    q_\zeta=F_\zeta(y),
    \qquad
    b_\zeta=\frac{q_\zeta+\mathbf v^\star_\zeta}{2},
    \label{eq:app_teacher_notation}
\end{equation}
where $\mathbf v^\star_\zeta$ is the ground-truth action-flow target.
The current robot state is included in the conditioning of $F_\zeta$.
All derivatives with respect to the future image hold these conditioning variables fixed.
For a quantity $u_\zeta$, we use
$\|u\|_\zeta=(\mathbb E_\zeta\|u_\zeta\|_2^2)^{1/2}$
to denote its root-mean-square magnitude over the shared noising conditions.

The analysis first treats the squared objectives in
Section~\ref{sec:video_action_loss}. Averaging the per-transition
expressions over training transitions gives the full training objective.
We discuss fixed weights and robust penalties at the end of this appendix.

\paragraph{Proposition 1: the combined supervision target.}
The sum of the matching and GT objectives satisfies
\begin{equation}
\begin{aligned}
    \mathcal L_{\mathrm{aux}}(\hat y)
    &=
    \mathbb E_\zeta
    \left[
        \|F_\zeta(\hat y)-q_\zeta\|_2^2
        +
        \|F_\zeta(\hat y)-\mathbf v^\star_\zeta\|_2^2
    \right]
    \\
    &=
    2\|F(\hat y)-b\|_\zeta^2
    +
    \frac{1}{2}\epsilon_{\mathrm{AoE}}^2,
    \qquad
    \epsilon_{\mathrm{AoE}}
    =
    \|q-\mathbf v^\star\|_\zeta .
    \label{eq:app_completed_square}
\end{aligned}
\end{equation}
Here $\epsilon_{\mathrm{AoE}}$ measures the frozen teacher's
action-flow error on the recorded transition.

\textit{Proof.}
For any vectors $u,q,v$,
\[
    \|u-q\|_2^2+\|u-v\|_2^2
    =
    2\left\|u-\frac{q+v}{2}\right\|_2^2
    +\frac{1}{2}\|q-v\|_2^2.
\]
Substituting the teacher predictions and taking expectations gives
Equation~\eqref{eq:app_completed_square}.
The final term is independent of the world-model parameters because
AoE, the recorded observation, and the ground-truth action target are fixed.

This identity also gives
\begin{equation}
\begin{aligned}
    \|F(\hat y)-q\|_\zeta
    &\le
    \|F(\hat y)-b\|_\zeta
    +\frac{\epsilon_{\mathrm{AoE}}}{2}
    \\
    &=
    \sqrt{
        \frac{\mathcal L_{\mathrm{aux}}(\hat y)}{2}
        -\frac{\epsilon_{\mathrm{AoE}}^2}{4}
    }
    +\frac{\epsilon_{\mathrm{AoE}}}{2}
    \\
    &\le
    \sqrt{\frac{\mathcal L_{\mathrm{aux}}(\hat y)}{2}}
    +\frac{\epsilon_{\mathrm{AoE}}}{2}.
    \label{eq:app_teacher_match_bound}
\end{aligned}
\end{equation}
The same bound holds for
$\|F(\hat y)-\mathbf v^\star\|_\zeta$.
Both follow from the triangle inequality for root-mean-square norms.
Thus the generated transition's action-flow error depends on its fit
to the auxiliary target and on the teacher's accuracy on the recorded
transition.

\paragraph{Proposition 2: local weighting of visual errors.}
Suppose $F_\zeta$ is twice differentiable near $y$, with uniformly
bounded first and second derivatives. Let
$J_\zeta(y)=\partial F_\zeta(y)/\partial y$.
For a sufficiently small image perturbation $\Delta y$,
\begin{equation}
    \mathcal L_{\mathrm{match}}(y+\Delta y)
    =
    \Delta y^\top
    \mathbb E_\zeta
    \left[J_\zeta(y)^\top J_\zeta(y)\right]
    \Delta y
    +O(\|\Delta y\|_2^3).
    \label{eq:app_image_geometry}
\end{equation}

\textit{Proof.}
Taylor expansion gives
\[
    F_\zeta(y+\Delta y)-F_\zeta(y)
    =
    J_\zeta(y)\Delta y+R_\zeta,
    \qquad
    \|R_\zeta\|_2\le c\|\Delta y\|_2^2
\]
for a constant $c$ in the local neighborhood.
Expanding the squared norm yields the quadratic term in
Equation~\eqref{eq:app_image_geometry}, a cross term of order
$O(\|\Delta y\|_2^3)$, and a remainder of order
$O(\|\Delta y\|_2^4)$.
Taking expectations establishes the result.

The matrix
$\mathbb E_\zeta[J_\zeta^\top J_\zeta]$
measures how visual changes affect the teacher's action predictions.
A visual change receives a larger penalty when it causes a larger
change in those predictions.
Its off-diagonal entries also account for interactions between pixels.
The attention visualizations in Appendix~A provide a complementary
view of the image regions used by AoE; the weighting in
Equation~\eqref{eq:app_image_geometry} is determined by its input
derivatives.

\paragraph{Proposition 3: propagation into world-model features.}
Let $h=H_\theta(\cdot)$ denote world-model features and let
$\hat y=G_\omega(h)$ denote the predicted future image.
We distinguish shared parameters $\theta$ from parameters $\omega$
specific to the video prediction path.
With the teacher fixed,
\begin{equation}
\begin{aligned}
    \nabla_h\mathcal L_{\mathrm{aux}}
    &=
    4\mathbb E_\zeta
    \left[
        \left(\frac{\partial \hat y}{\partial h}\right)^\top
        J_\zeta(\hat y)^\top
        \bigl(F_\zeta(\hat y)-b_\zeta\bigr)
    \right],
    \nabla_\theta\mathcal L_{\mathrm{aux}}
    =
    \left(\frac{\partial h}{\partial\theta}\right)^\top
    \nabla_h\mathcal L_{\mathrm{aux}}.
    \label{eq:app_shared_gradient}
\end{aligned}
\end{equation}

\textit{Proof.}
Differentiate Equation~\eqref{eq:app_completed_square} first with
respect to the teacher output, then through the teacher, the image
prediction path, and the shared feature extractor.
The constant teacher-error term has zero derivative.
This gives Equation~\eqref{eq:app_shared_gradient}.

Freezing AoE fixes its parameters while retaining the derivative of
its output with respect to its input image.
Consequently, the auxiliary objective provides a gradient to every
trainable component on the generated-image path.
The feature derivative can include all shared tensors used by the
action head, with those tensors collected into $h$.

For completeness, suppose the auxiliary objective has an
$L$-Lipschitz gradient in the local parameter neighborhood.
An auxiliary step with
$g=\nabla_\theta\mathcal L_{\mathrm{aux}}$
then satisfies
\begin{equation}
    \mathcal L_{\mathrm{aux}}(\theta-\eta g)
    \le
    \mathcal L_{\mathrm{aux}}(\theta)
    -
    \eta\left(1-\frac{L\eta}{2}\right)\|g\|_2^2.
    \label{eq:app_auxiliary_descent}
\end{equation}
This follows by integrating the gradient along the update segment
and applying its Lipschitz bound.
The auxiliary component therefore decreases its objective for
$0<\eta<2/L$ whenever $g\ne 0$.
Under joint training, the corresponding first-order change also
includes the inner product between the native and auxiliary gradients.

\paragraph{Propagation to native action predictions.}
Let $\hat{\mathbf a}_{\theta,\psi}=A_\psi(H_\theta(\cdot))$
be the native action prediction, with $\psi$ denoting parameters
specific to the action head.
The current observation, instruction, robot state, and native
sampling inputs are held fixed.
The auxiliary objective depends on the shared features and video
prediction path, so
$\nabla_\psi\mathcal L_{\mathrm{aux}}=0$.
Its contribution to the shared-parameter update changes the native
prediction according to
\begin{equation}
\begin{aligned}
    \hat{\mathbf a}_{\theta-\eta g,\psi}
    -
    \hat{\mathbf a}_{\theta,\psi}
    &=
    -\eta
    \frac{\partial\hat{\mathbf a}_{\theta,\psi}}{\partial\theta}
    g
    +O(\eta^2)
    \\
    &=
    -\eta
    \frac{\partial A_\psi}{\partial h}
    \frac{\partial h}{\partial\theta}
    \left(\frac{\partial h}{\partial\theta}\right)^\top
    \nabla_h\mathcal L_{\mathrm{aux}}
    +O(\eta^2).
    \label{eq:app_native_prediction_update}
\end{aligned}
\end{equation}
The first equality is a Taylor expansion in $\theta$; the second
uses Equation~\eqref{eq:app_shared_gradient}.
This expression isolates the auxiliary contribution to a parameter
update. Contributions from the native objective can be added at
first order.

The product
$(\partial h/\partial\theta)(\partial h/\partial\theta)^\top$
accounts for how parameter updates change the shared features.
It also makes explicit that optimizing model parameters and directly
optimizing a free feature vector generally produce different updates.

\paragraph{Conditions for transferring an action-error bound.}
Choose a reference feature vector $h^\star$ whose decoded future
satisfies $G_\omega(h^\star)=y$.
Define
\[
    g_\zeta(h)=F_\zeta(G_\omega(h)),
    \qquad
    J_\zeta^\star
    =
    \left.\frac{\partial g_\zeta}{\partial h}\right|_{h^\star},
    \qquad
    B
    =
    \left.\frac{\partial A_\psi}{\partial h}\right|_{h^\star}.
\]
We assume the teacher composition and native action map have
bounded second derivatives in the neighborhood under consideration.

The transfer condition requires a finite $\kappa$ such that
\begin{equation}
    \|B\Delta h\|_2
    \le
    \kappa
    \left(
        \mathbb E_\zeta
        \|J_\zeta^\star\Delta h\|_2^2
    \right)^{1/2}
    \label{eq:app_sensitivity_condition}
\end{equation}
for the local feature directions being analyzed.
This condition says that AoE responds to the feature changes that
affect the native action prediction.
It permits different output dimensions and coordinate conventions
for AoE and the native action head.

An equivalent condition over the full feature space is
$\ker M\subseteq\ker B$, where
\[
    M=\mathbb E_\zeta
    \left[(J_\zeta^\star)^\top J_\zeta^\star\right].
\]
To verify the equivalence, first observe that
Equation~\eqref{eq:app_sensitivity_condition} forces $B\Delta h=0$
whenever $M\Delta h=0$.
Conversely, if $\ker M\subseteq\ker B$, then
\[
    B
    =
    B(M^\dagger)^{1/2}M^{1/2},
\]
where $M^\dagger$ inverts the positive eigenvalues of $M$ and is
zero on its null space.
Taking norms proves Equation~\eqref{eq:app_sensitivity_condition}
with
$\kappa=\|B(M^\dagger)^{1/2}\|_{\mathrm{op}}$.

\paragraph{Proposition 4: a local bound on native action error.}
Let $\mathbf a^\star$ be the ground-truth target for the native
action prediction, and define its reference error as
\[
    \epsilon_{\mathrm{head}}
    =
    \|A_\psi(h^\star)-\mathbf a^\star\|_2.
\]
Under the conditions above, there exists a local constant $C_h$
such that
\begin{equation}
    \|A_\psi(h)-\mathbf a^\star\|_2
    \le
    \epsilon_{\mathrm{head}}
    +
    \kappa
    \left(
        \sqrt{\frac{\mathcal L_{\mathrm{aux}}(h)}{2}}
        +
        \frac{\epsilon_{\mathrm{AoE}}}{2}
    \right)
    +
    C_h\|h-h^\star\|_2^2.
    \label{eq:app_native_error_bound}
\end{equation}

\textit{Proof.}
Write $\Delta h=h-h^\star$.
By Taylor expansion,
\begin{equation}
\begin{aligned}
    A_\psi(h)-A_\psi(h^\star)
    &=B\Delta h+R_a,
    &
    \|R_a\|_2
    &\le c_a\|\Delta h\|_2^2,
    \\
    g_\zeta(h)-g_\zeta(h^\star)
    &=J_\zeta^\star\Delta h+R_{g,\zeta},
    &
    \|R_g\|_\zeta
    &\le c_g\|\Delta h\|_2^2.
\end{aligned}
\end{equation}
The sensitivity condition and the triangle inequality give
\begin{equation}
\begin{aligned}
    \|A_\psi(h)-A_\psi(h^\star)\|_2
    &\le
    \kappa\|J^\star\Delta h\|_\zeta
    +c_a\|\Delta h\|_2^2
    \\
    &\le
    \kappa\|g(h)-g(h^\star)\|_\zeta
    +(c_a+\kappa c_g)\|\Delta h\|_2^2.
\end{aligned}
\end{equation}
Since $G_\omega(h^\star)=y$, we have $g_\zeta(h^\star)=q_\zeta$.
Applying Equation~\eqref{eq:app_teacher_match_bound} and adding
$\epsilon_{\mathrm{head}}$ proves
Equation~\eqref{eq:app_native_error_bound}, with
$C_h=c_a+\kappa c_g$.

The result connects three quantities: the generated future's fit
to the auxiliary target, AoE's accuracy on recorded transitions,
and the native head's accuracy at the reference features.
For fixed reference quantities and local constants, reducing the
auxiliary loss tightens the action-error bound.

\paragraph{Imperfect reference reconstruction and partial coverage.}
The reference features can also decode an approximation to the
recorded future.
Define
\[
    \epsilon_{\mathrm{ref}}
    =
    \|g(h^\star)-q\|_\zeta.
\]
The same proof adds $\kappa\epsilon_{\mathrm{ref}}$ to
Equation~\eqref{eq:app_native_error_bound}, because
\[
    \|g(h)-g(h^\star)\|_\zeta
    \le
    \|g(h)-q\|_\zeta+\epsilon_{\mathrm{ref}}.
\]

A corresponding extension covers feature directions that AoE
does not measure.
Let $P=M^\dagger M$ project onto the directions with nonzero
teacher sensitivity.
Decomposing $\Delta h$ gives
\begin{equation}
    \|B\Delta h\|_2
    \le
    \kappa\|J^\star\Delta h\|_\zeta
    +
    \|B(I-P)\Delta h\|_2.
\end{equation}
Thus the general bound includes
$\|B(I-P)(h-h^\star)\|_2$ as an additional term.
The auxiliary objective constrains the measured directions, while
the native objective also trains action-relevant directions outside
that set.

\paragraph{How the action head continues to learn.}
Shared-feature updates also change the training inputs and gradients
of the native action head.
Let
$g_\psi(\theta,\psi)=\nabla_\psi\mathcal L_{\mathrm{native}}(\theta,\psi)$.
For an auxiliary change
$\Delta\theta=-\eta\nabla_\theta\mathcal L_{\mathrm{aux}}$,
differentiability gives
\begin{equation}
\begin{aligned}
    g_\psi(\theta+\Delta\theta,\psi)
    &=
    g_\psi(\theta,\psi)
    +
    \nabla^2_{\psi\theta}\mathcal L_{\mathrm{native}}
    \,\Delta\theta
    +o(\|\Delta\theta\|_2)
    \\
    &=
    g_\psi(\theta,\psi)
    -
    \eta
    \nabla^2_{\psi\theta}\mathcal L_{\mathrm{native}}
    \nabla_\theta\mathcal L_{\mathrm{aux}}
    +o(\eta).
    \label{eq:app_indirect_head_learning}
\end{aligned}
\end{equation}
This is the first-order expansion of the native head's gradient.
It describes the auxiliary contribution to subsequent native
updates: the world model supplies modified features, and the
action head learns from those features through its original
training objective.

\paragraph{From action-flow discrepancies to sampled actions.}
The preceding analysis applies to the native prediction map used
in the loss. For a flow-based head, a bound on velocity prediction
can be propagated through its sampler under trajectory coverage.

To make this step explicit for AoE, let $u_{\hat y}(\tau)$ and
$u_y(\tau)$ be action trajectories generated under the predicted
and recorded futures, starting from the same noise.
Suppose the generated-conditioned velocity field is
$L_f$-Lipschitz in the action input.
Subtracting the two flow equations yields
\[
    \frac{d}{d\tau}
    \|u_{\hat y}(\tau)-u_y(\tau)\|_2
    \le
    L_f\|u_{\hat y}(\tau)-u_y(\tau)\|_2
    +e(\tau),
\]
where
\[
    e(\tau)
    =
    \left\|
        f_\phi(o_t,\hat y;u_y(\tau),\tau)
        -
        f_\phi(o_t,y;u_y(\tau),\tau)
    \right\|_2.
\]
Multiplying by $e^{-L_f\tau}$ and integrating gives
\begin{equation}
    \|u_{\hat y}(1)-u_y(1)\|_2
    \le
    \int_0^1 e^{L_f(1-\tau)}e(\tau)\,d\tau.
    \label{eq:app_flow_execution_bound}
\end{equation}
If the reference inference-trajectory distribution has density
at most $C_{\mathrm{cov}}$ times the training noising distribution,
then the Cauchy--Schwarz inequality gives
\begin{equation}
    \mathbb E
    \|u_{\hat y}(1)-u_y(1)\|_2^2
    \le
    C_{\mathrm{cov}}
    \left(\int_0^1 e^{2L_f(1-\tau)}\,d\tau\right)
    \mathcal L_{\mathrm{match}}.
\end{equation}
The density condition allows the velocity error along inference
trajectories to be controlled by the training expectation.
The same argument applies to a native flow head once its velocity
discrepancy is bounded.
Task success is measured through the execution evaluations in
Section~4.

\paragraph{Fixed weights and robust implementations.}
For positive scalar weights $\lambda_m$ and $\lambda_g$, the
combined squared objective has the same form with
\begin{equation}
\begin{aligned}
    b_\zeta^{(\lambda)}
    &=
    \frac{\lambda_m q_\zeta+
          \lambda_g\mathbf v^\star_\zeta}
         {\lambda_m+\lambda_g},
    \\
    \lambda_m\mathcal L_{\mathrm{match}}
    +\lambda_g\mathcal L_{\mathrm{GT}}
    &=
    (\lambda_m+\lambda_g)
    \|F(\hat y)-b^{(\lambda)}\|_\zeta^2
    +
    \frac{\lambda_m\lambda_g}
         {\lambda_m+\lambda_g}
    \epsilon_{\mathrm{AoE}}^2.
\end{aligned}
\end{equation}
This follows by the same square expansion as Proposition~1.

Fixed coordinate masks and reliability weights restrict the
analysis to the supervised coordinates and replace the Euclidean
norm with the corresponding weighted norm.
For a smooth robust penalty $\rho$, differentiation produces the
same teacher-to-image gradient path, with the residual replaced
by its weighted derivative $\rho'$.
Near zero residual, the matching objective has local quadratic
form
$J^\top WJ$, where $W$ contains the coordinate weights and the
curvature of $\rho$ at zero.

For the penalty
$\rho(r)=\sqrt{1+r^2}-1$ used in the implementation,
\[
    \rho(r)=\frac{r^2}{\sqrt{1+r^2}+1}.
\]
Therefore, on a bounded residual range $|r|\le R$,
\[
    \frac{r^2}{\sqrt{1+R^2}+1}
    \le \rho(r)\le \frac{r^2}{2}.
\]
With positive weights on the supervised coordinates, the robust
and squared objectives consequently bound one another locally.
The squared-loss results then carry over with modified constants
on those coordinates.

\section{Training and Evaluation Details}
\label{app:training_details}

This appendix summarizes the data sources, pretraining configuration,
world-model settings, replay protocol, and downstream WAM training
configurations used in our experiments. Unless otherwise specified, all
settings are kept fixed across the corresponding comparisons.

We separate training provenance from evaluation protocol because the two
serve different purposes in the study. The former determines what information
a model can use to learn a representation, while the latter determines
whether the same representation transfers when task identities, object
arrangements, or rendering conditions change. Each table therefore records the
source data and optimization choices first, followed by the protocol that is
held fixed in the corresponding comparison.

The same convention applies to AoE. A checkpoint is trained once in the
canonical-action stage and then reused as a frozen inverse-dynamics model in
evaluation and VITA training. Freezing the checkpoint makes the teacher
interface identical across world models and WAMs, so differences between rows
can be attributed to the image source or the auxiliary objective rather than
to a changing action translator. Whenever a table reports replay, the replay
environment and initial state are synchronized with the image pair supplied
to AoE.

\subsection{Canonical-Action Pretraining}
\label{app:canonical_action_pretraining}

AoE is initialized through canonical-action pretraining on a large-scale
robot corpus. The corpus contains $9{,}637$ hours of robot
interaction data. The authoritative corpus total is reported in
Table~\ref{tab:mae_pretrain_data}; Table 9 reports the duration and source type of each constituent dataset.

The canonical-action model is trained on 64 H800 GPUs using BF16
precision. We use a global batch size of 1,024 and train for 600K
optimization steps. Episodes are sampled with a mixing temperature of
1.0, and each episode provides 1,536 image pairs for canonical-action
learning. The remaining optimization and regularization settings are
listed in Table~\ref{tab:mae_pretrain_setting}. Validation is performed
on held-out data from DROID, AgibotWorld, and LET.

These validation streams are used for checkpoint selection and calibration
only; they do not provide examples from the held-out downstream tasks used in
the cross-task tests. The pretraining stage therefore supplies a common
camera-space action interface before any target benchmark is introduced.
Each training pair is formed from observations that share the same episode
context but are separated by a valid interaction interval. Pair sampling
preserves the endpoint relation that AoE must recover, while state dropout
prevents the model from relying on a single proprioceptive cue when the visual
transition already determines the motion. The resulting checkpoint is
exported once and reused without parameter updates in all replay, VITA, and
WAM comparisons.

\begin{table*}[t]
\centering
\caption{\textbf{Data sources for AoE canonical-action pretraining.}
The reported hours are taken directly from the training configuration.}
\label{tab:mae_pretrain_data}
\renewcommand{\arraystretch}{1.1}
\resizebox{0.5\linewidth}{!}{
\begin{tabular}{l | c c}
\toprule
\textbf{Data source} & \textbf{Duration (h)} & \textbf{Type} \\
\midrule
DROID-all-aligned       & 164.0   & Real-robot \\
AgibotWorld-Beta-all    & 2174.0  & Real-robot \\
LET-all-aligned         & 171.0   & Real-robot \\
Xingzhi                 & 260.0   & Real-robot \\
Bodeng-Xinghai          & 1431.0  & Real-robot \\
Songling                & 957.0   & Real-robot \\
EgoDex                  & 722.0   & Human ego \\
Groot-live              & 8.0     & Simulation \\
CALVIN                  & 38.4    & Simulation \\
RMBench                 & 64.5    & Simulation \\
RoboCasa                & 122.1   & Simulation \\
Intern-A1               & 3525.0  & Simulation \\
\midrule
\textbf{Overall}        & 9637    & --- \\
\bottomrule
\end{tabular}
}
\end{table*}

\begin{table*}[t]
\centering
\caption{\textbf{AoE canonical-action pretraining configuration.}}
\label{tab:mae_pretrain_setting}
\renewcommand{\arraystretch}{1.15}
\resizebox{0.55\linewidth}{!}{
\begin{tabular}{l l}
\toprule
\rowcolor{gray!8}
\textbf{Category} & \textbf{Setting} \\
\midrule
Hardware & 64 H800 GPUs (8 nodes $\times$ 8 GPUs) \\
Precision & BF16 \\
Per-GPU batch size & 16 \\
Global batch size & 1024 \\
Training steps & 600K \\
Warmup steps & 3K \\
Optimizer learning rate & $2\times10^{-4}$ \\
Shuffle buffer & 16,384 \\
Pairs sampled per episode & 1,536 \\
Dataset mixing temperature & 1.0 \\
State dropout & 0.5 \\
Visual encoder & Independent DINO \\
Validation coverage & DROID, AgibotWorld, and LET \\
\bottomrule
\end{tabular}
}
\end{table*}

The data table records corpus provenance, whereas the configuration table
records the optimization contract. Keeping these views separate is useful
when reproducing the study: changing the mixture changes the teacher's
transfer behavior, while changing the optimizer or sampling schedule changes
the quality of the same camera-space target. All downstream rows use the
resulting checkpoint under the same frozen-parameter convention.

\subsection{World-Model Training and Replay}
\label{app:world_model_training}

We compare two Cosmos training configurations. Cosmos + VITA uses the
frozen AoE inverse-dynamics model as an action-grounded supervisor and
optimizes the VITA objective. Cosmos follows the same model, data,
and temporal configuration without IDM supervision or auxiliary VITA
losses. The optimizer learning rates follow their respective training
configurations. The complete settings are listed in
Table~\ref{tab:world_model_setting}.

The comparison is made at the level of the supervision path: both models
receive the same visual context, temporal window, resolution, hardware
budget, and number of optimization iterations, while only the AoE-guided
terms are enabled for Cosmos + VITA. This matched setup is important because
the replay experiment evaluates the future images produced by the model,
rather than a separately trained action policy.

The VITA branch is evaluated only on transitions for which the required image
pair and teacher output are available. It does not alter the recorded target
video, the native diffusion objective, or the sampling procedure used to
produce candidates at test time. At inference, candidates are sampled from
the corresponding world-model checkpoint and passed through the same frozen
AoE and simulator replay code used for the plain Cosmos condition.

For Cosmos + VITA, the frozen AoE model provides action-grounded
supervision while its parameters remain unchanged. The matching,
ground-truth, and ROI terms are combined using the coefficients shown in
the table. Cosmos does not use any of these terms.

\begin{table*}[t]
\centering
\caption{\textbf{World-model training and replay settings.}
Cosmos + VITA and Cosmos share the same architecture, data, and
temporal configuration. Their optimizer learning rates follow the
corresponding configurations shown below. Cosmos + VITA additionally uses
the frozen AoE teacher and the VITA objective, whereas Cosmos is
trained without IDM supervision.}
\label{tab:world_model_setting}
\renewcommand{\arraystretch}{1.15}
\resizebox{0.75\linewidth}{!}{
\begin{tabular}{l l l}
\toprule
\rowcolor{gray!8}
\textbf{Component} & \textbf{Cosmos + VITA} & \textbf{Cosmos} \\
\midrule
Hardware
& 4 nodes $\times$ 8 H800 GPUs
& 4 nodes $\times$ 8 H800 GPUs
\\
IDM supervision
& Frozen AoE teacher
& None
\\
Dataloader batch size
& 4
& 4
\\
Video resolution
& $432\times768$
& $432\times768$
\\
Temporal window
& 93 continuous frames
& 93 continuous frames
\\
Window stride
& 92
& 92
\\
Optimizer learning rate
& $2^{-15}=3.0518\times10^{-5}$
& $2^{-15}=3.0518\times10^{-5}$
\\
Warmup steps
& 1,000
& 1,000
\\
Maximum iterations
& 10,000
& 10,000
\\
$\lambda_{\mathrm{match}}$
& 1.0
& Not applicable
\\
$\lambda_{\mathrm{GT}}$
& 1.0
& Not applicable
\\
$\lambda_{\mathrm{ROI}}$
& 0.1
& Not applicable
\\
\bottomrule
\end{tabular}
}
\end{table*}

The table should be read row-wise: the first block defines the common
world-model training contract, and the last three rows specify the terms that
exist only in the VITA condition. The replay evaluation never fine-tunes AoE
on generated frames. Consequently, an image that receives a high execution
score must contain enough endpoint and interaction information for the same
pretrained translator that handled recorded images.

For RoboTwin, the 50 tasks are divided into two groups of 25 tasks.
We use two cross-task folds, with the training and held-out groups
exchanged between folds. Each task is evaluated under both clean and
randomized configurations, resulting in 100 task--configuration pairs.
For each replay condition, AoE predicts actions from image pairs and the
predicted actions are converted into RoboTwin joint-position commands
for simulator replay. The replay IDM checkpoint is frozen and receives
no additional training during evaluation. The complete replay protocol
is summarized in Table~\ref{tab:world_model_replay_protocol}.

\begin{table*}[t]
\centering
\caption{\textbf{Cross-task world-model replay protocol on RoboTwin.}
The 50 tasks are divided into two groups of 25 tasks for two-fold
cross-task evaluation. Each task is evaluated under clean and
randomized configurations.}
\label{tab:world_model_replay_protocol}
\renewcommand{\arraystretch}{1.15}
\resizebox{0.92\linewidth}{!}{
\begin{tabular}{l l}
\toprule
\rowcolor{gray!8}
\textbf{Category} & \textbf{Setting} \\
\midrule
Task split
& 25 training tasks and 25 held-out tasks
\\
Cross-validation
& Two folds, with the two task groups exchanged between training and evaluation
\\
Number of task--configuration pairs
& 100
\\
Replay procedure
& AoE predicts actions from image pairs, followed by RoboTwin joint-position replay
\\
IDM replay checkpoint
& No additional training for the replay IDM
\\
\bottomrule
\end{tabular}
}
\end{table*}

The replay rows use a shared reset state for all candidates in a pass@k
measurement. This removes the possibility that a successful candidate gains
an easier initial configuration than the other candidates. Clean and
randomized configurations are rendered with the same task identity, so the
cross-task split tests task transfer while the configuration axis tests
robustness to scene variation. The detailed task membership is listed later
in Table~\ref{tab:robotwin_25_25_split}.

\subsection{Downstream WAM Training}
\label{app:wam_training}

We apply the same frozen AoE teacher to two WAM architectures,
LingBot-VA and FastWAM. AoE parameters remain frozen throughout all
downstream experiments. The VITA objective is evaluated on both LIBERO
and RoboTwin, with the model-specific temporal and input configurations
listed in Table~\ref{tab:wam_training_setting}.

For both LingBot-VA and FastWAM, the LIBERO and RoboTwin models are trained for 5,000 and 50,000 optimization steps, respectively. Table~\ref{tab:wam_training_setting} summarizes the hardware, batch sizes, gradient accumulation, optimizer settings, and VITA-specific hyperparameters used in these experiments.

In all WAM experiments, the VITA loss is added during training, while each model retains its native action-generation pipeline for evaluation, including preprocessing, normalization, action decoding, and closed-loop replanning.

Across both architectures and benchmarks, AoE remains frozen, and the matching, GT, and ROI loss coefficients are fixed at 1.0, 1.0, and 0.1, respectively. Batch sizes and optimizer settings are specified separately for each architecture in Table~13. This keeps the auxiliary supervision consistent while accommodating the training configuration of each WAM.

\begin{table}[H]
\centering
\caption{\textbf{WAM training configurations.} AoE is frozen in downstream WAM training. RoboTwin uses clean-only training for C $\to$ C+R and combined clean/randomized training for C+R $\to$ C+R.}
\label{tab:wam_training_setting}
\renewcommand{\arraystretch}{1.15}
\resizebox{0.7\linewidth}{!}{
\begin{tabular}{l | c c || c c}
\toprule
& \multicolumn{2}{c||}{\textbf{LingBot-VA}} & \multicolumn{2}{c}{\textbf{FastWAM}} \\
\cmidrule(lr){2-3}\cmidrule(lr){4-5}
\rowcolor{gray!8}
\textbf{Setting} & \textbf{LIBERO} & \textbf{RoboTwin} & \textbf{LIBERO} & \textbf{RoboTwin} \\
\midrule
Hardware & 32 H800 & 32 H800 & 32 H800 & 32 H800 \\
Micro batch & 4 & 4 & 8 & 8 \\
Gradient accumulation & 2 & 2 & 2 & 2 \\
Global batch & 256 & 256 & 512 & 512 \\
Training steps & 5,000 & 50,000 & 5,000 & 50,000\\
Learning rate & $10^{-5}$ & $10^{-5}$ & $10^{-4}$ & $10^{-4}$ \\
Weight decay & 0.1 & 0.1 & 0.01 & 0.01 \\
Warmup steps & 500 & 3,000 & 500 & 3,000 \\
AoE & Frozen & Frozen & Frozen & Frozen \\
VITA chunk size & 8 & 8 & 8 & 8 \\
VITA interval & 2 & 2 & 1 & 1 \\
$\lambda_{\mathrm{match}}$ & 1.0 & 1.0 & 1.0 & 1.0 \\
$\lambda_{\mathrm{GT}}$ & 1.0 & 1.0 & 1.0 & 1.0 \\
$\lambda_{\mathrm{ROI}}$ & 0.1 & 0.1 & 0.1 & 0.1 \\
\bottomrule
\end{tabular}
}
\end{table}

The two WAM architectures expose different native temporal interfaces, so we
report their settings in a single table rather than forcing a common sequence
length or action frequency. The shared quantities are the frozen teacher and
the VITA coefficients; the model-specific quantities preserve each
implementation's original preprocessing and temporal alignment. This is why
the same downstream comparison can be read across LIBERO and RoboTwin without
interpreting a change in input format as a method change.

\subsection{RoboTwin Task Partition}
\label{app:robotwin_task_partition}

Table~\ref{tab:robotwin_25_25_split} lists the complete RoboTwin
25/25 partition. The partition is fixed before training and is shared
by the world-model replay and native WAM transfer experiments. In
Fold~1, Group~1 is used for training and Group~2 is held out for
evaluation. Fold~2 reverses the two groups. This procedure ensures
that each task group serves as the held-out task set once.

\begin{table*}[t]
\centering
\caption{\textbf{RoboTwin 25/25 task split used for cross-task training and evaluation.}
The two folds exchange the training and held-out groups: Fold~1 trains
on Group~1 and evaluates on Group~2, while Fold~2 reverses the two
groups. The same split is used for world-model replay and native WAM
transfer.}
\label{tab:robotwin_25_25_split}
\renewcommand{\arraystretch}{1.08}
\resizebox{0.5\linewidth}{!}{
\begin{tabular}{c l | c l}
\toprule
\rowcolor{gray!8}
\textbf{ID} & \textbf{Group 1}
& \textbf{ID} & \textbf{Group 2} \\
\midrule
1  & adjust\_bottle
   & 1  & blocks\_ranking\_size \\
2  & beat\_block\_hammer
   & 2  & dump\_bin\_bigbin \\
3  & blocks\_ranking\_rgb
   & 3  & handover\_mic \\
4  & click\_alarmclock
   & 4  & lift\_pot \\
5  & move\_can\_pot
   & 5  & click\_bell \\
6  & grab\_roller
   & 6  & move\_pillbottle\_pad \\
7  & handover\_block
   & 7  & move\_playingcard\_away \\
8  & hanging\_mug
   & 8  & move\_stapler\_pad \\
9  & open\_laptop
   & 9  & place\_a2b\_right \\
10 & open\_microwave
   & 10 & place\_bread\_skillet \\
11 & pick\_diverse\_bottles
   & 11 & place\_burger\_fries \\
12 & pick\_dual\_bottles
   & 12 & place\_cans\_plasticbox \\
13 & place\_a2b\_left
   & 13 & place\_empty\_cup \\
14 & place\_bread\_basket
   & 14 & place\_fan \\
15 & place\_can\_basket
   & 15 & place\_mouse\_pad \\
16 & place\_container\_plate
   & 16 & place\_object\_basket \\
17 & place\_dual\_shoes
   & 17 & place\_object\_scale \\
18 & place\_object\_stand
   & 18 & place\_phone\_stand \\
19 & press\_stapler
   & 19 & place\_shoe \\
20 & rotate\_qrcode
   & 20 & put\_bottles\_dustbin \\
21 & scan\_object
   & 21 & put\_object\_cabinet \\
22 & shake\_bottle
   & 22 & shake\_bottle\_horizontally \\
23 & stack\_blocks\_two
   & 23 & stack\_blocks\_three \\
24 & stamp\_seal
   & 24 & stack\_bowls\_three \\
25 & stack\_bowls\_two
   & 25 & turn\_switch \\
\bottomrule
\end{tabular}
}
\end{table*}

No task is moved between groups after a model has been trained, and
configuration randomization is applied after the task-level assignment. Thus,
the held-out fold measures transfer to task identities that are absent from
training, while the clean/randomized pairing measures robustness within each
identity. Reporting both folds gives every task group one training role and
one held-out role.

\section{Detailed experimental results}
\label{app:full_results}

\subsection{Paired execution calibration}
\begin{table}[H]
\centering
\caption{\textbf{Calibration of execution and image recovery.} The same 300 instances, with successful recorded-action replay and visible grippers, are used in all rows; they cover 60.0\% of 500 candidates. Independent geometric recovery reaches 92.0\% on real futures.}
\label{tab:calibration}
\renewcommand{\arraystretch}{1.15}
\resizebox{0.52\linewidth}{!}{
\begin{tabular}{l | c}
\toprule
\rowcolor{gray!8}
\textbf{Input and recovery path} & \textbf{Success (\%)} \\
\midrule
Dataset action replay & \tb{100.0} \\
GT camera actions & \ul{98.0} \\
\rowcolor{blue!4}
Dataset image replay + AoE & 93.7 \\
VAE-reconstructed dataset image + AoE & 91.3 \\
Generated image + AoE & 2.0 \\
Generated image + geometry & 3.0 \\
\bottomrule
\end{tabular}
}
\end{table}

The paired set supports comparison of the camera-space adapter, image decoder, VAE reconstruction, and generated-future recovery. The full-population LIBERO table remains the overall performance reference. Calibration coverage and success are separate quantities: the former describes the tested operating range, and the latter describes behavior within it.

\subsection{Visual-source diagnosis, human judgments, and local interventions}
\label{app:visual_diagnosis}
\begin{figure}[H]
\centering\includegraphics[width=\linewidth]{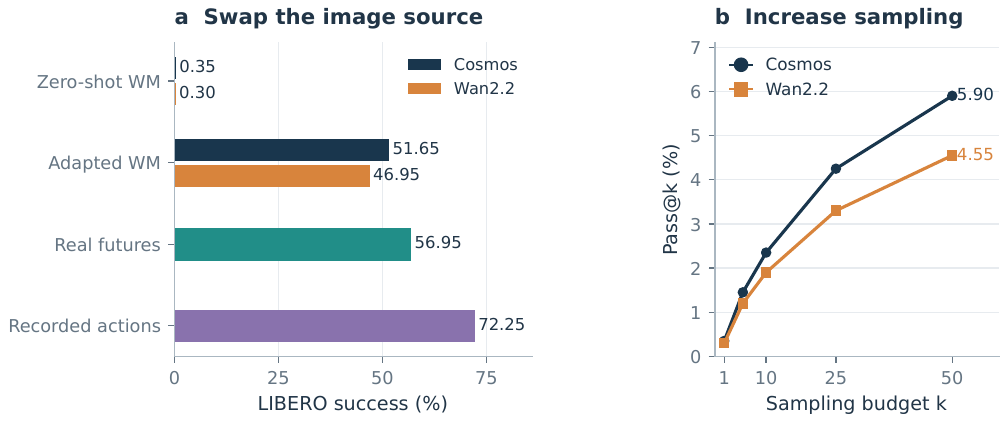}
\caption{\textbf{Visual source and sampling budget.} Left: episode-weighted LIBERO success with a shared pretrained AoE for image sources; ``Adapted WM'' denotes the VITA-supervised world model. Right: any-success@$k$ with all candidate executions starting from a common reset state.}
\label{fig:diagnosis}
\end{figure}

A blinded study samples 200 rollouts from each of five future sources and assigns three independent raters to every rollout. Raters see the initial observation, instruction, and video, with model identity and execution outcome hidden. They judge visual plausibility and apparent task completion. Majority vote defines each binary label. Display size, frame rate, and clip duration are shared across sources. Execution is evaluated on the same examples.

\begin{figure}[H]
\centering\includegraphics[width=\linewidth]{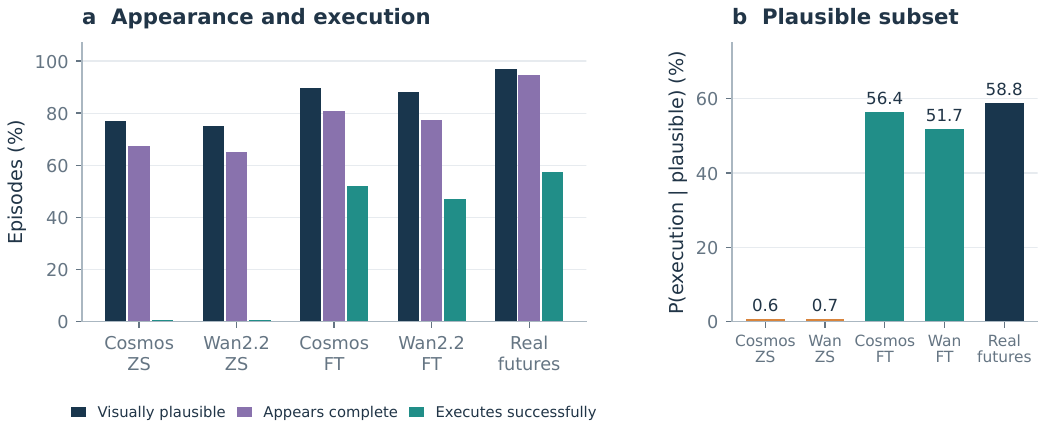}
\caption{\textbf{Visual plausibility and recovered execution.} Human judgments and simulator outcomes on 200 videos per source, with three blinded raters per video. The right panel reports execution conditional on a plausible judgment; ``FT'' denotes the VITA-supervised condition.}
\label{fig:perception}
\end{figure}

\label{sec:human}
\begin{table}[H]
\centering
\caption{\textbf{Paired human judgments and execution.} Each source contains 200 videos judged by three raters. P denotes majority-rated plausibility. All numeric entries are percentages; this subset is separate from the full benchmark evaluations.}
\label{tab:human_full}
\renewcommand{\arraystretch}{1.15}
\resizebox{\linewidth}{!}{
\begin{tabular}{l | c c c c c c c}
\toprule
\rowcolor{gray!8}
\textbf{Image source} & \textbf{Source} & \textbf{Inference} & \textbf{Plausible} & \textbf{Looks done} & \textbf{Execute} & \textbf{Execute$\mid$P} & \textbf{Looks done$\mid$Fail} \\
\midrule
\rowcolor{blue!4}
Dataset image & Replay & -- & 97.0 & 94.5 & 57.5 & 58.8 & 90.6 \\
\midrule
Cosmos & Zero-shot & pass@1 & 77.0 & 67.5 & 0.5 & 0.6 & 67.3 \\
WAN2.2 & Zero-shot & pass@1 & 75.0 & 65.0 & 0.5 & 0.7 & 64.8 \\
\rowcolor{blue!4}
Cosmos + VITA & Zero-shot & pass@1 & 89.5 & 81.0 & 52.0 & 56.4 & 65.6 \\
\rowcolor{blue!4}
WAN2.2 + VITA & Zero-shot & pass@1 & 88.0 & 77.5 & 47.0 & 51.7 & 61.3 \\
\bottomrule
\end{tabular}
}
\end{table}

\paragraph{Local geometry and execution tolerance.}
\label{sec:intervention}
We use 200 short aligned segments to vary gripper--object geometry by 0, 1, 2, 4, and 8 pixels. Background interventions match the edit area and processing steps. Independent raters score plausibility; action recovery and execution share the original paired reset state. A four-pixel interaction offset reduces execution from 75.5\% to 23.0\%, while 86.0\% of videos retain a plausible judgment. The matched background edit retains 72.5\% execution. In a complementary experiment, repairing generated segments gives 3.0\% success for background repair, 18.5\% for gripper repair, and 32.0\% for interaction-region repair, compared with 1.5\% before repair. Repairs use aligned ground-truth regions in these short segments.

\begin{figure}[H]
\centering\includegraphics[width=\linewidth]{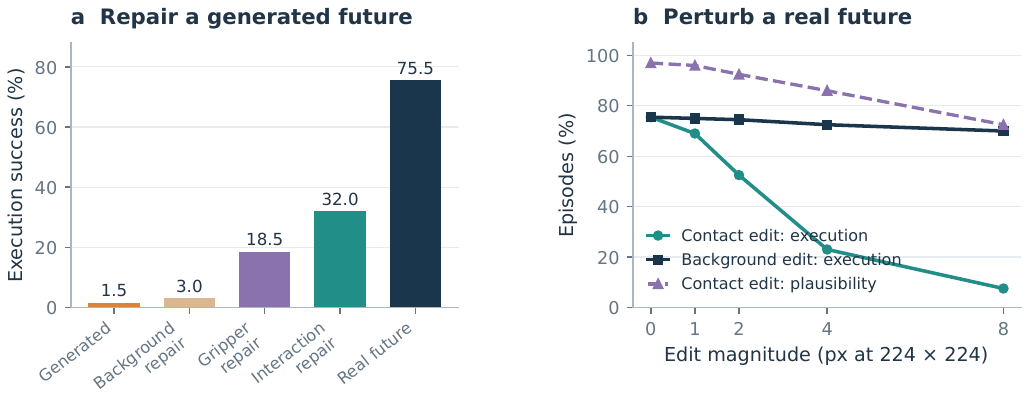}
\caption{\textbf{Local image edits and recovered control.} Left: execution after repairing selected regions of generated futures. Right: execution after controlled interaction-region or background offsets in recorded futures, together with human-rated plausibility for interaction edits.}
\label{fig:intervention}
\end{figure}

\paragraph{Offline metric comparison.}
For an offline metric comparison, every evaluated pair has a matched reference future. We orient each score so that a larger value predicts better action recovery. SSIM~\citep{ssim} and DINO compare the paired images; LPIPS~\citep{lpips} is negated. AoE action agreement is the negative normalized discrepancy between actions recovered from predicted and reference futures. Geometric consistency compares independently recovered end-effector and interaction geometry. Scores are evaluated within tasks before aggregation. The comparison uses held-out episodes for evaluation and a separate source-task partition for selecting score normalization. FVD~\citep{fvd} is a distribution-level quantity and is recorded separately at the model/checkpoint level.

\begin{table}[H]
\centering
\caption{\textbf{Offline scores versus execution.} AUROC and rank correlation are computed on matched image pairs and aggregated within tasks.}
\label{tab:metric}
\renewcommand{\arraystretch}{1.15}
\resizebox{0.62\linewidth}{!}{
\begin{tabular}{l | c c}
\toprule
\rowcolor{gray!8}
\textbf{Offline score} & \textbf{Within-task AUROC} & \textbf{Spearman $\rho$} \\
\midrule
SSIM & 0.55 & 0.08 \\
DINO feature similarity & 0.61 & 0.17 \\
Negative LPIPS & 0.60 & 0.15 \\
\rowcolor{blue!4}
AoE action agreement & \ul{0.79} & \ul{0.49} \\
Geometric consistency & \tb{0.82} & \tb{0.55} \\
\bottomrule
\end{tabular}
}
\end{table}

The paired score comparison shows why generic image similarity is not used as
the sole selection criterion. SSIM, DINO similarity, and LPIPS summarize
appearance agreement, whereas AoE action agreement and geometric consistency
use the same endpoint-to-action interface that is ultimately executed. The
ranking is computed inside each task before aggregation to prevent tasks with
more episodes from dominating the correlation statistics.

\subsection{Candidate and target-data budgets}
\begin{table}[H]
\centering
\caption{\textbf{Candidate budget on LIBERO.} Success (\%) over 2,000 initial states, with candidates executed from a shared reset state. Two decimal places retain the precision of this sampling sweep.}
\label{tab:candidate_budget}
\renewcommand{\arraystretch}{1.15}
\resizebox{0.5\linewidth}{!}{
\begin{tabular}{l | c c c c}
\toprule
\rowcolor{gray!8}
\textbf{Model} & \textbf{IDM action} & \textbf{Source} & \textbf{Inference} & \textbf{Success (\%)} \\
\midrule
\multirow{5}{*}{Cosmos + AoE} & \checkmark & Zero-shot & pass@1 & 0.35 \\
 & \checkmark & Zero-shot & pass@5 & 1.45 \\
 & \checkmark & Zero-shot & pass@10 & 2.35 \\
 & \checkmark & Zero-shot & pass@25 & 4.25 \\
 & \checkmark & Zero-shot & pass@50 & 5.90 \\
 \midrule
\multirow{5}{*}{WAN2.2 + AoE} & \checkmark & Zero-shot & pass@1 & 0.30 \\
 & \checkmark & Zero-shot & pass@5 & 1.20 \\
 & \checkmark & Zero-shot & pass@10 & 1.90 \\
 & \checkmark & Zero-shot & pass@25 & 3.30 \\
 & \checkmark & Zero-shot & pass@50 & 4.55 \\
\bottomrule
\end{tabular}
}
\end{table}

\begin{table}[H]
\centering
\caption{\textbf{Target-task adaptation budget.} Adaptation begins from the same-domain source-task checkpoint and uses 500 common evaluation episodes. This target-data sweep is separate from cross-task VITA training. The dataset-image reference is 57.2\%.}
\label{tab:target_budget}
\renewcommand{\arraystretch}{1.15}
\resizebox{0.5\linewidth}{!}{
\begin{tabular}{l | c c c}
\toprule
\rowcolor{gray!8}
\textbf{Model} & \textbf{Target demonstrations} & \textbf{Inference} & \textbf{Success (\%)} \\
\midrule
\multirow{6}{*}{Cosmos} & 0 & pass@1 & 3.6 \\
 & 5 & pass@1 & 18.2 \\
 & 10 & pass@1 & 29.6 \\
 & 25 & pass@1 & 43.4 \\
 & 50 & pass@1 & 54.2 \\
 & 100 & pass@1 & 55.6 \\
 \midrule
\multirow{6}{*}{WAN2.2} & 0 & pass@1 & 3.0 \\
 & 5 & pass@1 & 15.0 \\
 & 10 & pass@1 & 25.2 \\
 & 25 & pass@1 & 38.6 \\
 & 50 & pass@1 & 50.0 \\
 & 100 & pass@1 & 53.0 \\
\bottomrule
\end{tabular}
}
\end{table}
The candidate sweep and target-data sweep answer different questions. The
first measures how much can be gained by drawing more futures from one
zero-shot checkpoint under a fixed reset. The second measures how quickly a
small amount of target-task adaptation closes the gap to the recorded-image
reference. Keeping these tables separate avoids treating test-time sampling
and training-time adaptation as interchangeable sources of information.

\subsection{Supervision objective ablations}
\label{app:ablation}
\begin{table}[H]
\centering
\caption{\textbf{Supervision ablation on FastWAM.} Held-out success uses LIBERO-90 training and 2,000 LIBERO test episodes; LIBERO-Plus evaluates environment shifts. Native baseline and full-VITA Overall values follow the main-text table.}
\label{tab:fastwam_ablation}
\renewcommand{\arraystretch}{1.15}
\resizebox{0.7\linewidth}{!}{
\begin{tabular}{l | c c}
\toprule
\rowcolor{gray!8}
\textbf{Training objective} & \textbf{Held-out tasks (\%)} & \textbf{LIBERO-Plus (\%)} \\
\midrule
Native baseline & 11.2 & 51.5 \\
Matched-budget native & 11.5 & 50.9 \\
Future DINO alignment & 11.8 & 54.0 \\
AoE smoothness & 14.5 & 51.5 \\
\rowcolor{blue!4}
VITA (Match + GT + ROI) & \ul{23.6} & \ul{55.9} \\
Match + GT & \tb{24.1} & \tb{56.4} \\
\bottomrule
\end{tabular}
}
\end{table}

The matched-budget native row allocates the additional training compute to the original losses. Future DINO alignment uses the same observed futures and a frozen semantic teacher. The AoE smoothness row uses recovered motion regularity. The full auxiliary objective combines paired teacher matching, confidence-weighted GT flow supervision, and a gripper-centered ROI reconstruction loss. The Match + GT row has zero ROI weight and reaches 24.1\% held-out success, compared with 23.6\% for the full objective. This comparison measures the contribution of the local reconstruction term alongside action-flow supervision. Full loss definitions appear in Appendix~\ref{app:loss_details}. All comparisons use the same source-task partition, frozen pretrained AoE, and native inference path.

The matched-budget row controls for extra optimization, while Future DINO
alignment controls for adding a semantic image target without an executable
action interface. AoE smoothness controls for a regularizer derived from the
teacher without the paired future-image target. Together, these rows isolate
the contribution of the action-grounded supervision used by VITA.

\subsection{Native WAM control on RoboTwin}
\begin{table}[H]
\centering
\caption{\textbf{RoboTwin performance with native WAM inference.} C and R denote clean and randomized configurations. Values (\%) match the corresponding in-distribution and out-of-distribution main-text tables.}
\label{tab:robotwin_native}
\renewcommand{\arraystretch}{1.15}
\resizebox{0.7\linewidth}{!}{
\begin{tabular}{l | c c c || c c}
\toprule
& \multicolumn{3}{c||}{\textbf{RoboTwin: C+R $\rightarrow$ C+R}}
& \multicolumn{2}{c}{\textbf{RoboTwin: C $\rightarrow$ C+R}} \\
\cmidrule(lr){2-4}\cmidrule(lr){5-6}
\rowcolor{gray!8}
\textbf{Method} & \textbf{Easy} & \textbf{Hard} & \textbf{Avg.} & \textbf{Easy} & \textbf{Hard} \\
\midrule
FastWAM & 86.4 & 88.2 & 87.3 & 69.5 & 9.3 \\
\rowcolor{blue!4}
FastWAM + VITA & \ul{90.7} & \tb{91.9} & \ul{91.3} & 73.8 & 17.5 \\
LingBot-VA & 85.6 & 84.2 & 84.9 & \ul{74.3} & \ul{19.2} \\
\rowcolor{blue!4}
LingBot-VA + VITA & \tb{92.4} & \ul{91.2} & \tb{91.8} & \tb{82.5} & \tb{31.6} \\
\bottomrule
\end{tabular}
}
\end{table}

The native WAM table reports the same action decoder and closed-loop evaluator
for every row. The first block measures training and testing on the combined
clean/randomized distribution, while the second block trains on clean scenes
and tests on both clean and randomized scenes. The second block therefore
combines task transfer with appearance and layout changes, making the larger
VITA gain a direct measure of robustness under a shifted evaluation condition.

\section{Teacher-independent evaluation of VITA-supervised futures}
\label{app:teacher_independence}

The main evaluation uses AoE to both supervise the VITA objective and recover
actions from generated futures. This appendix tests whether the resulting
future videos remain executable when the action-recovery model is changed
after training. We keep the WAN2.2 + VITA checkpoint, initial observations,
instructions, generated videos, simulator states, and execution code fixed;
only the IDM used to translate each future video into an action is replaced.
This isolates the dependence of the measured success rate on the evaluator
rather than on a change in the world-model sample.

We use J-IDM from VERA~\citep{vera} and the inverse-dynamics model used in
DreamGen~\citep{dreamgen}. The ``IDM zero-shot'' column records whether the
IDM has been evaluated on a task distribution unseen during its own training.
AoE is marked with a checkmark because its canonical-action pretraining is
performed independently of the held-out LIBERO-Plus and RoboTwin tasks. The
J-IDM and DreamGen-IDM rows are marked with a cross: both models are trained
in-distribution for this evaluation and are not used as cross-task teachers.
Their scores are therefore teacher-independence controls, while AoE remains
the zero-shot cross-task reference.

\begin{table*}[t]
\centering
\caption{\textbf{Teacher-independent evaluation of a VITA-supervised WAN2.2.}
All rows use the same WAN2.2 checkpoint trained with the AoE-based VITA loss;
only the IDM used to recover executable actions at evaluation is changed.
The checkmark in ``IDM zero-shot'' denotes evaluation on tasks unseen during
IDM training. J-IDM~\citep{vera} and DreamGen-IDM~\citep{dreamgen} are
in-distribution controls and therefore are marked with a cross. Results are
success rates in percent.}
\label{tab:teacher_independence}
\renewcommand{\arraystretch}{1.15}
\resizebox{\linewidth}{!}{
\begin{tabular}{l | c || c c c c c c c | c || c c}
\toprule
& & \multicolumn{8}{c||}{\textbf{LIBERO-Plus}} &
\multicolumn{2}{c}{\textbf{RoboTwin}} \\
\cmidrule(lr){3-10}
\cmidrule(lr){11-12}
\textbf{IDM evaluator} & \textbf{IDM zero-shot}
& \textbf{Background} & \textbf{Light} & \textbf{Language}
& \textbf{Noise} & \textbf{Layout} & \textbf{Camera}
& \textbf{Robot Init.} & \textbf{Overall}
& \textbf{Easy} & \textbf{Hard} \\
\midrule
\rowcolor{blue!4}
WAN2.2 + AoE
& \checkmark
& \ul{38.5} & \ul{45.3} & \ul{37.1} & \ul{0.1}
& \tb{39.7} & 26.6 & 13.4 & \ul{29.9}
& 59.6 & 40.4 \\
WAN2.2 + J-IDM
& $\times$
& 19.5 & 31.6 & \tb{42.8} & 0.0
& \ul{39.2} & \tb{27.7} & -- & 26.8
& \tb{65.5} & \tb{66.7} \\
WAN2.2 + DreamGen-IDM
& $\times$
& \tb{39.4} & \tb{46.1} & 36.2 & \tb{9.4}
& 35.2 & \ul{26.1} & -- & \tb{30.8}
& \ul{61.9} & \ul{57.8} \\
\bottomrule
\end{tabular}
}
\end{table*}

\paragraph{Recorded-video replay.}
We first evaluate action recovery from recorded dataset videos to provide
a reference for each IDM. Table~\ref{tab:teacher_independence_replay}
reports these results alongside recorded-action replay. Over the six
evaluated LIBERO-Plus categories, AoE, J-IDM, and DreamGen-IDM achieve
47.0\%, 55.5\%, and 54.8\% success, respectively, compared with 76.2\%
for recorded-action replay. These results establish that all three IDMs
can recover executable actions from recorded visual transitions, while
also revealing differences in their recovery performance across
perturbations. In particular, sensor noise substantially degrades image-based
action recovery for all three IDMs. Recorded-video replay therefore
provides a decoder-specific reference rather than assuming that action
recovery is equally reliable across all evaluation conditions.

\begin{table*}[t]
\centering
\caption{\textbf{Recorded-video replay with different IDM evaluators
on LIBERO-Plus.}
The image-replay rows recover actions from recorded dataset videos;
the action-replay row directly executes recorded actions as an oracle
reference. Results are success rates in percent.
Overall is computed by pooling successful episodes across the six
evaluated categories, totaling 8,480 episodes, rather than averaging
category-level percentages. Robot Init.\ is not evaluated in this
replay comparison.}
\label{tab:teacher_independence_replay}
\renewcommand{\arraystretch}{1.15}
\resizebox{\linewidth}{!}{
\begin{tabular}{l | c c c c c c c | c}
\toprule
\textbf{Replay setting}
& \textbf{Background} & \textbf{Light} & \textbf{Language}
& \textbf{Noise} & \textbf{Layout} & \textbf{Camera}
& \textbf{Robot Init.} & \textbf{Overall} \\
\midrule
Dataset action replay (Oracle)
& 71.9 & 75.3 & 78.0 & 83.8 & 78.8 & 67.7 & -- & 76.2 \\
\midrule
\rowcolor{blue!4}
Dataset image replay + AoE
& 60.7 & 60.3 & 59.6 & 0.0 & 53.6 & 57.0 & -- & 47.0 \\
Dataset image replay + J-IDM
& 65.6 & 74.5 & 66.1 & 0.9 & 60.1 & 75.4 & -- & 55.5 \\
Dataset image replay + DreamGen-IDM
& 66.4 & 80.0 & 71.9 & 3.6 & 67.2 & 52.2 & -- & 54.8 \\
\bottomrule
\end{tabular}
}
\end{table*}

\paragraph{Generated futures without VITA.}
We next evaluate WAN2.2 futures generated without VITA supervision
using the same three IDM evaluators.
Table~\ref{tab:teacher_independence_without_vita} shows that these
futures yield low execution success with every decoder:
0.5\% with AoE, 1.8\% with J-IDM, and 0.9\% with DreamGen-IDM.
Together with the recorded-video results, this comparison shows that
the execution gap between recorded and unsupervised generated futures
persists beyond the AoE interface.

\begin{table*}[t]
\centering
\caption{\textbf{Teacher-independent evaluation of WAN2.2 futures
without VITA supervision on LIBERO-Plus.}
All rows evaluate futures generated without the VITA loss.
AoE is the zero-shot cross-task IDM, while J-IDM and DreamGen-IDM
are the in-distribution controls described in
Table~\ref{tab:teacher_independence}.
Results are success rates in percent.
Overall pools successful episodes over the evaluated categories:
10,030 episodes across seven categories for AoE, and 8,480 episodes
across six categories for J-IDM and DreamGen-IDM.
The latter two exclude Robot Init.; their Overall values therefore
cover a different evaluation set from AoE's.}
\label{tab:teacher_independence_without_vita}
\renewcommand{\arraystretch}{1.15}
\resizebox{\linewidth}{!}{
\begin{tabular}{l | c || c c c c c c c | c}
\toprule
\textbf{IDM evaluator} & \textbf{IDM zero-shot}
& \textbf{Background} & \textbf{Light} & \textbf{Language}
& \textbf{Noise} & \textbf{Layout} & \textbf{Camera}
& \textbf{Robot Init.} & \textbf{Overall} \\
\midrule
\rowcolor{blue!4}
WAN2.2 + AoE
& $\checkmark$
& 0.9 & 0.5 & 0.5 & 0.0 & 0.5 & 0.7 & 0.3 & 0.5 \\
WAN2.2 + J-IDM
& $\times$
& 5.5 & 1.4 & 2.4 & 0.4 & 1.1 & 0.9 & -- & 1.8 \\
WAN2.2 + DreamGen-IDM
& $\times$
& 2.7 & 0.7 & 0.9 & 0.1 & 0.5 & 1.2 & -- & 0.9 \\
\bottomrule
\end{tabular}
}
\end{table*}

\paragraph{VITA gains across IDM evaluators.}
Comparing Table~\ref{tab:teacher_independence_without_vita}
with Table~\ref{tab:teacher_independence}, VITA increases
LIBERO-Plus execution success from 0.5\% to 29.9\% with AoE,
from 1.8\% to 26.8\% with J-IDM, and from 0.9\% to 30.8\%
with DreamGen-IDM. These correspond to improvements of approximately
29.4, 25.0, and 29.9 percentage points, respectively.
We compare each evaluator against its own no-VITA baseline;
the Overall values should not be interpreted as a ranking across
IDMs because their category coverage differs.
The improvements with J-IDM and DreamGen-IDM show that the benefit
of VITA is not restricted to action recovery by its AoE teacher.
Nevertheless, VITA-supervised futures remain below the recorded-video
replay reference for each evaluator, leaving a residual gap between
generated and recorded visual transitions.

The RoboTwin results in Table~\ref{tab:teacher_independence}
provide complementary evidence that VITA-supervised futures remain
executable under alternative action-recovery interfaces.
J-IDM achieves 65.5\%/66.7\% on Easy/Hard, and DreamGen-IDM
achieves 61.9\%/57.8\%, compared with 59.6\%/40.4\% for AoE.
Since J-IDM and DreamGen-IDM are trained in-distribution for these
evaluations, their results should not be interpreted as zero-shot
cross-task generalization by the IDMs themselves. Rather, the
LIBERO-Plus comparisons demonstrate improvements under independently
trained evaluators, while the RoboTwin results confirm that the
VITA-supervised futures support execution beyond the teacher's own
action-recovery interface. Together, these controls provide evidence
that VITA improves executable visual information rather than merely
matching AoE-specific responses.

\section{Complete training objective and gradient implementation}
\label{app:loss_details}

We detail the implemented LingBot-VA objective and the auxiliary terms in Eq.~\eqref{eq:aux_loss}. Native losses preserve each host model's training conventions. Batch indices are suppressed in the auxiliary equations; each per-frame loss is averaged over the minibatch.

\paragraph{1. Native video flow matching.}
For clean video latent $z_0$ and Gaussian noise $\epsilon_z$, LingBot-VA constructs
\begin{equation}
 z_\sigma=(1-\sigma)z_0+\sigma\epsilon_z,\qquad
 v_z^*=\epsilon_z-z_0,\qquad
 \hat v_z=f_\theta(z_\sigma,\sigma,\mathrm{text},\mathrm{history}).
\end{equation}
Each latent frame independently samples its diffusion timestep. With $B$ sequences, $F$ latent frames, and $N_z$ spatial-channel entries per frame,
\begin{equation}
 \mathcal L_{\mathrm{video}}=
 \frac1{BF}\sum_{b,f}w_z(\sigma_{b,f})
 \frac{\|\hat v_{z,b,f}-v_{z,b,f}^*\|_2^2}{N_z}.
\end{equation}
The scheduler supplies $w_z$; spatial and channel dimensions are averaged before batch and time.

\paragraph{2. Native action flow matching.}
The native action tensor follows its corresponding clean-to-noise path,
\begin{align}
 a_{\sigma_a}&=(1-\sigma_a)a_0+\sigma_a\epsilon_a,
 &v_a^*&=\epsilon_a-a_0,\\
 \mathcal L_{\mathrm{native\text{-}action}}
 &=\frac1{BF}\sum_{b,f}w_a(\sigma_{a,b,f})
 \frac{\sum_d m_{b,f,d}(\hat v_{a,b,f,d}-v_{a,b,f,d}^*)^2}
 {\sum_d m_{b,f,d}+\varepsilon_{\mathrm{den}}}.
\end{align}
LingBot-VA retains its 30-channel action tensor. LIBERO supervises the first seven channels, with channel seven using $-1$ for open and $+1$ for close. The canonical AoE target supplies the auxiliary video losses. FastWAM retains its own native action representation, scheduler, and loss definitions.

\paragraph{3. Canonical action and coordinate weights.}
For each future time $t$, $A_t\in\mathbb R^{2\times12}$ contains, per arm, $[u_0,v_0,u_t,v_t,\Delta z_t,R^{6D}_{0\rightarrow t},g_t]$, with normalized absolute UV coordinates and normalized depth displacement. The start-to-future relative rotation occupies six coordinates. Expanding the five group weights $[0,1,1,1,1]$ gives
\begin{equation}
 G=(0,0,1,1,1,1,1,1,1,1,1,1).
\end{equation}
These coordinate weights apply to Match and GT supervision; the complete action tensor remains available in the noisy action input. The arm mask $M_{t,a}$ expresses visibility and label validity for GT supervision and ground-truth gripper ROI construction.

\paragraph{4. Shared AoE action-flow samples.}
AoE uses $A_{\tau,t}=(1-\tau_t)\xi_t+\tau_t A_t$ and target $V_t^*=A_t-\xi_t$, with uniform $\tau_t$ and Gaussian $\xi_t$. Both branches in Eq.~\eqref{eq:teacher_branches} share the same $A_{\tau,t}$, $\tau_t$, and $\xi_t$. Their common conditions are the window's first image and robot state.

\begin{equation}
 P_t=T_\phi(\hat I_t;c_t),\qquad
 R_t=\operatorname{sg}[T_\phi(I_t^{\mathrm{GT}};c_t)],
 \label{eq:teacher_branches}
\end{equation}
where $c_t$ collects the shared initial image, robot state, noisy action, and flow time.

\paragraph{5. Match loss.}
Define the weighted robust reduction
\begin{align}
 \mathcal H_w(X,Y)&=
 \frac{\sum_{a,d}w_{t,a,d}\rho_\delta((X_{t,a,d}-Y_{t,a,d})/s_d)}
 {\sum_{a,d}w_{t,a,d}+\varepsilon_{\mathrm{den}}},\nonumber\\
 \rho_\delta(r)&=\delta^2\!\left(\sqrt{1+(r/\delta)^2}-1\right).
 \label{eq:robust_reduction}
\end{align}
All coordinate scales satisfy $s_d=1$, and $\delta=1$. Match uses $w_{t,a,d}=G_dC^{\mathrm{off}}_{t,a,d}$ and compares the predicted-video response $P_t$ with the detached genuine-video response $R_t$. It is computed whenever the matching term is well-defined, including when the gripper leaves the field of view, and can therefore provide complementary supervision when the corresponding GT supervision is unavailable.

\paragraph{6. GT flow loss and online confidence.}
GT compares $P_t$ with the analytic velocity $V_t^*$ using weights $MG C^{\mathrm{off}}C^{\mathrm{on}}$. Online confidence is computed from the genuine-video branch:
\begin{align}
 e_{t,a,d}&=\operatorname{clip}\!\left[
 \left(\frac{R_{t,a,d}-V_{t,a,d}^*}{\eta_d}\right)^2,0,z_{\max}^2\right],\\
 C^{\mathrm{on}}_{t,a,d}&=\operatorname{clip}\!\left[
 \exp\!\left(-\frac{e_{t,a,d}+|\mathcal G(d)|^{-1}
 \sum_{j\in\mathcal G(d)}e_{t,a,j}}{4}\right),c_{\min},1\right].
\end{align}
Here $\mathcal G(d)$ is the action group containing coordinate $d$, $c_{\min}=0.05$, and $z_{\max}=4$. Tolerances are $0.05$ for each start/end UV coordinate, $0.10$ for depth, $0.20$ for each rotation coordinate, and $0.25$ for grip. Confidence is detached and weights supervision according to teacher reliability on the genuine pair.

\paragraph{7. Offline confidence.}
Calibration cells index data domain, action-flow timestep bin, arm, and coordinate. The five bins are $[0,0.2)$, $[0.2,0.4)$, $[0.4,0.6)$, $[0.6,0.8)$, and $[0.8,1]$. For calibration residual $r$ and temperature $T_{\mathrm{cal}}$, the cell's basic confidence is
\begin{equation}
 \widetilde C^{\mathrm{off}}_{c}
 =\exp\!\left[-\widehat{\mathbb E}_{c}[r^2]/T_{\mathrm{cal}}\right].
\end{equation}
Cell sample counts determine shrinkage toward a global conservative fallback. Unavailable domain--timestep cells use that fallback. The resulting table is frozen during WAM training.

\paragraph{8. Ground-truth gripper ROI reconstruction.}
Let $p_{t,a}$ be the ground-truth future gripper projection in pixel coordinates. We form
\begin{align}
 Q_{t,a}(x,y)&=M_{t,a}\exp\!\left[-\frac{\|(x,y)-p_{t,a}\|_2^2}{2\sigma_{\mathrm{px}}^2}\right],
 &Q_t(x,y)&=\max_a Q_{t,a}(x,y),\\
 W_t(x,y)&=\frac{1+\alpha Q_t(x,y)}
 {\operatorname{mean}_{x,y}[1+\alpha Q_t(x,y)]},
 &\alpha&=4,\quad\sigma_{\mathrm{px}}=10.
\end{align}
The normalized weights emphasize the GT gripper endpoints. With $C$ color channels, reconstruction uses
\begin{equation}
 \mathcal L_{\mathrm{ROI},t}
 =\frac{\sum_{c,x,y}W_t(x,y)
 \sqrt{(\hat I_{t,c,x,y}-I^{\mathrm{GT}}_{t,c,x,y})^2+\epsilon_{\mathrm{pix}}^2}}
 {C\sum_{x,y}W_t(x,y)},\qquad\epsilon_{\mathrm{pix}}=10^{-3}.
\end{equation}
The implemented ROI uses these gripper-endpoint Gaussians. The spatial weighting is computed entirely from ground-truth projections and arm validity.

\paragraph{9. All-future-frame objective.}
Equation~\eqref{eq:aux_loss} averages Match, GT, and ROI over all $T=4(F-1)$ genuine future frames at stride one. Their weights are respectively $1$, $1$, and $0.1$. Teacher evaluation uses chunks of eight temporal pairs, retaining every pair and the same temporal average.

\begin{equation}
 \mathcal L_{\mathrm{aux}}=\frac1T\sum_{t=1}^{T}
 \left(\mathcal L_{\mathrm{match},t}+\mathcal L_{\mathrm{GT},t}
 +0.1\mathcal L_{\mathrm{ROI},t}\right).
 \label{eq:aux_loss}
\end{equation}

\paragraph{10. Optimizer-step schedule.}
For zero-based optimizer step $s$, let $\gamma(s)=\min(1,(s+1)/1000)$ and $K=2$. Define $\Pi_{\mathrm{vs}}$ as the gradient projection onto video-branch and shared parameters. The implemented update uses
\begin{equation}
 g_s=\nabla_\theta(\mathcal L_{\mathrm{video}}+\mathcal L_{\mathrm{native\text{-}action}})
 +\mathbf1[(s+1)\bmod K=0]\,K\gamma(s)\,
 \Pi_{\mathrm{vs}}\nabla_\theta\mathcal L_{\mathrm{aux}}.
 \label{eq:actual_gradient}
\end{equation}
Native gradients are evaluated every optimizer step. Each twice step evaluates the complete temporal auxiliary objective and scales its gradient by two, compensating for its evaluation frequency. Warmup reaches full auxiliary weight at the 1,000th optimizer step.

\paragraph{11. Frozen modules and gradient routing.}
LingBot-VA's native flow definition gives the clean-latent estimate $\hat z_0=z_\sigma-\sigma\hat v_z$. The differentiable chain is
\begin{equation}
 \hat v_z\longrightarrow\hat z_0\longrightarrow
 \hat I_t\longrightarrow T_\phi\longrightarrow\mathcal L_{\mathrm{aux}}.
\end{equation}
The ROI term also contributes a direct image-space gradient through $\hat I_t$, bypassing the teacher. AoE and the Wan VAE have frozen parameters. The genuine-video teacher branch is evaluated with gradients disabled. A VAE vector--Jacobian product propagates image-space derivatives to the predicted latents. Auxiliary gradients update video-branch and shared-transformer parameters; action-only parameters receive the native action-flow loss. These routes preserve the native video/action inputs, targets, and inference paths while adding action-aware supervision to video predictions.

\end{document}